\documentclass[10pt,journal,compsoc]{IEEEtran}
\usepackage{amsmath,amsfonts}
\usepackage{algorithmic}
\usepackage{algorithm}
\usepackage{array}
\usepackage[caption=false,font=normalsize,labelfont=sf,textfont=sf]{subfig}
\usepackage{textcomp}
\usepackage{stfloats}
\usepackage{url}
\usepackage{verbatim}
\usepackage[colorlinks,
            linkcolor=red,   
            anchorcolor=blue,  
            citecolor=green,        
            ]{hyperref}
\usepackage{graphicx}
\usepackage{cite}
\usepackage{times}
\usepackage{epsfig}
\usepackage{amssymb}
\usepackage{booktabs}
\usepackage{multirow}
\usepackage{flushend}
\usepackage{verbatim}
\usepackage{adjustbox}
\usepackage{pifont}
\usepackage{enumitem}
\usepackage{makecell}
\usepackage{comment}
\usepackage{array}
\usepackage{xcolor}         % colors
\usepackage{booktabs}
\usepackage{epigraph}
\usepackage{longtable}
\usepackage{threeparttable}
\usepackage{multirow}
\usepackage{ragged2e}

\begin{document}


% \markboth{Journal of \LaTeX\ Class Files,~Vol.~14, No.~8, August~2015}
\markboth{}%Preprint}%
{Shell \MakeLowercase{\textit{et al.}}: Bare Advanced Demo of IEEEtran.cls for IEEE Computer Society Journals}
\title{Towards Knowledge-driven Autonomous Driving}
\author{Xin Li$^{*}$,~ 
       Yeqi Bai$^{*}$,~ 
       Pinlong Cai$^{* \dagger}$,~ 
       Licheng Wen,~ 
       Daocheng Fu,~ 
       Bo Zhang,~ 
       Xuemeng Yang,~ 
       Xinyu Cai,~ \\ 
       Tao Ma,~  
       Jianfei Guo,~   
       Xing Gao,~   
       Min Dou,~  
       Yikang Li,~  
       Botian Shi$^\dagger$,~
       Yong Liu,~  
       Liang He~   
       and Yu~Qiao
\IEEEcompsocitemizethanks{

\IEEEcompsocthanksitem 
X. Li, Y. Bai, P. Cai,  L. Wen,  D. Fu, B. Zhang, X. Yang,  X. Cai, T. Ma, J. Guo, X. Gao, M. Dou, Y. Li, B. Shi and Y. Qiao are with Shanghai Artificial Intelligence Laboratory. 
\IEEEcompsocthanksitem 
X. Li and L. He are also with East China Normal University.
\IEEEcompsocthanksitem 
T. Ma is also with the Chinese University of Hong Kong.  
\IEEEcompsocthanksitem 
Y. Liu is with Zhejiang University.
\IEEEcompsocthanksitem $^*$ indicates equal contribution. $^{\dagger}$ denotes corresponding authors: Pinlong Cai ({caipinlong@pjlab.org.cn}) and Botian Shi ({shibotian@pjlab.org.cn})
}
        % <-this % stops a space
% \thanks{*Corresponding author}
\markboth{Preprint}%
{Shell \MakeLowercase{\textit{et al.}}: A Sample Article Using IEEEtran.cls for IEEE Journals}}
% \maketitle
{\fnsymbol{footnote}} 
{}

\IEEEtitleabstractindextext{%
\begin{abstract}
\justifying
This paper explores the emerging knowledge-driven autonomous driving technologies. Our investigation highlights the limitations of current autonomous driving systems, in particular their sensitivity to data bias, difficulty in handling long-tail scenarios, and lack of interpretability. Conversely, knowledge-driven methods with the abilities of cognition, generalization and life-long learning emerge as a promising way to overcome these challenges. This paper delves into the essence of knowledge-driven autonomous driving and examines its core components: dataset \& benchmark, environment, and driver agent. By leveraging large language models, world models, neural rendering, and other advanced artificial intelligence techniques, these components collectively contribute to a more holistic, adaptive, and intelligent autonomous driving system. The paper systematically organizes and reviews previous research efforts in this area, and provides insights and guidance for future research and practical applications of autonomous driving.
We will continually share the latest updates on cutting-edge developments in knowledge-driven autonomous driving along with the relevant valuable open-source resources at: \url{https://github.com/PJLab-ADG/awesome-knowledge-driven-AD}.
\end{abstract}

}

\maketitle

% is not selected <OR> if conference mode is selected - because compsoc
% conference papers position the abstract like regular (non-compsoc)
\IEEEdisplaynontitleabstractindextext

% \fi
\IEEEpeerreviewmaketitle

\ifCLASSOPTIONcompsoc
\IEEEraisesectionheading{
\label{sec:introduction}}
% \newpage
\section{Introduction}
\label{sec:intro}
\IEEEPARstart{I}n recent years, autonomous driving has undergone substantial development, primarily propelled by continuous advancements in sensor technology \cite{li2020lidar, van2018autonomous, xiang2023multi}, rapid progress in machine learning and artificial intelligence (AI) \cite{zhang2022beverse, hu2023planning, chen2022milestonessurvey}, as well as innovations in high-precision mapping and positioning technologies  \cite{bao2022high, cheng2022review}, etc. The positive influence of regulations and policies has further contributed to this progress. Despite noteworthy advancements in autonomous driving, persistent challenges remain. An overreliance on data-driven approaches exposes systems to data bias, resulting in overfitting on training data \cite{cao2022autonomous, huang2023sug}. This challenge impedes existing autonomous driving systems from effectively addressing long-tail and cross-domain issues \cite{xiang2023multi, wang2022parallel}, thereby limiting their adaptability in new environments. Moreover, existing autonomous driving systems lack interpretability \cite{zablocki2022explainability, fu2023drive, zhang2023dsiv}. Data-driven algorithms are often perceived as black boxes, presenting a challenge to provide human-understandable explanations for their decisions. This challenge impedes the ability to confirm whether the model genuinely makes intelligent decisions and restricts the potential for guiding further optimization of the system. Despite numerous attempts to address these issues \cite{shao2023safety, jing2022inaction, guan2022integrated}, no universally reliable method can satisfactorily resolve them. Consequently, addressing challenges such as data bias, long-tail issues, cross-domain problems, and the lack of interpretability remains a critical focus for ongoing research and development in autonomous driving.

Contemporary autonomous driving methodologies involve training models through the accumulation of extensive datasets to impart proficient driving capabilities \cite{yu2022autonomous, masello2022traditional}. Data-driven models tend to prioritize common cases while overlooking rare corner cases. This constraint is rooted in the assumption of independence and identical distribution (i.i.d.) underlying data-driven methodologies, which proves challenging to meet in real-world scenarios  \cite{wu2022trajectory, fantauzzo2022feddrive, chellapandi2023survey}.  Despite the expanding scale of data collection, the inherent limitation arises from the inadequacy of limited data to encompass an infinite array of corner cases \cite{fu2023drive, bogdoll2021description, liu2022curse}. To make fundamental strides in autonomous driving, it is crucial to explore technological changes and replicate human learning patterns in driving through modeling \cite{wang2022social, sestino2022let}. As underscored by Yann LeCun \cite{lecun2022path}, human proficiency in mastering fundamental driving skills and adeptly adapting to diverse and unpredictable scenarios, such as navigating complex traffic conditions and changing weather, requires merely dozens of hours of professional practice. This accentuates the efficient learning and knowledge summarization capabilities inherent in humans.

\begin{figure*}
    \centering
    \includegraphics[width=0.81\linewidth]{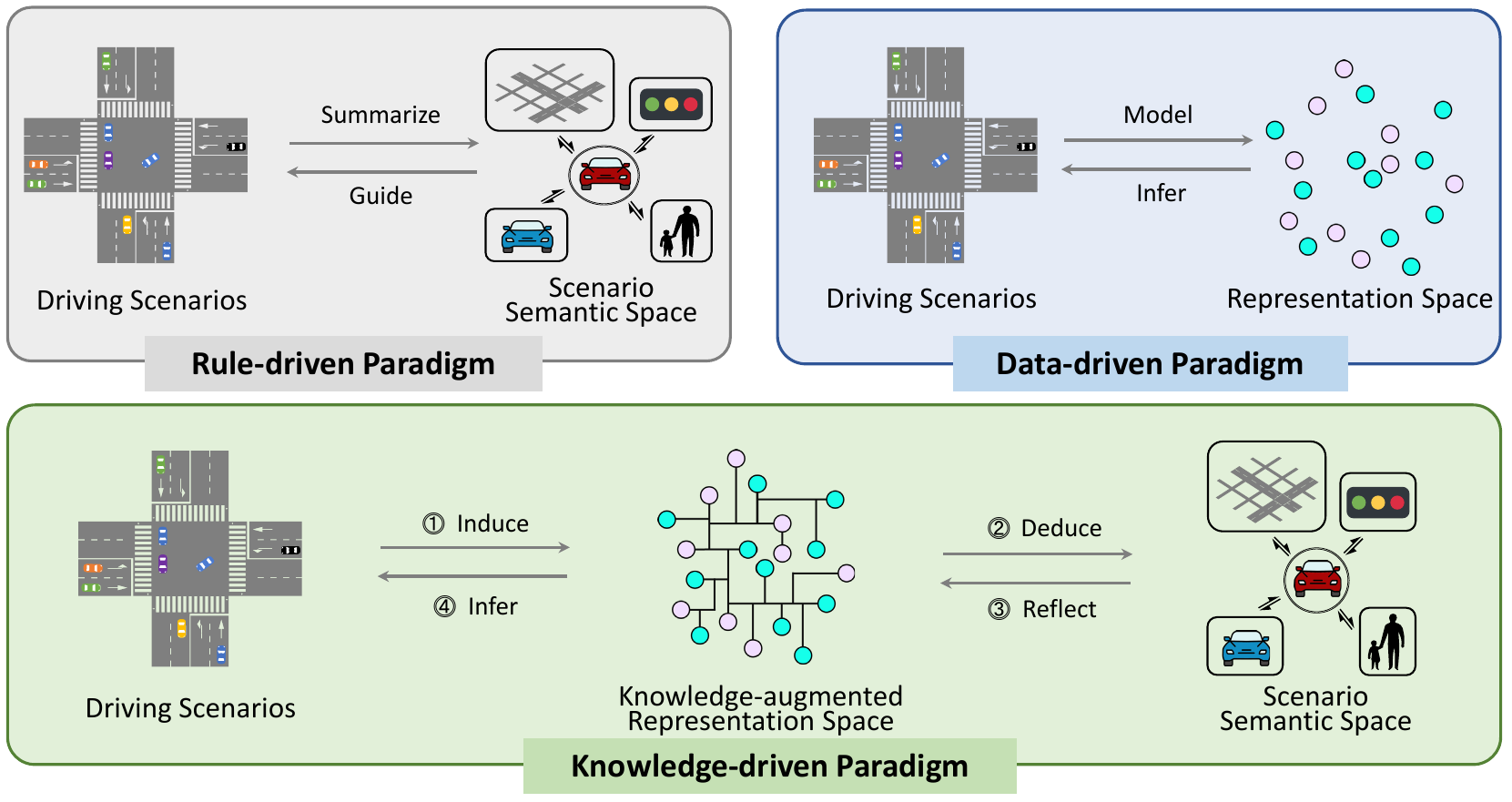}
    \caption{
    Comparison of three technical paradigms to autonomous driving. (1) The rule-based paradigm utilizes the understanding of driving scenarios that are summarized in the scenario semantic space to guide driving. (2) The data-based paradigm tends to model the driving scenarios into the representation space, which is subsequently inferred to the real world to accomplish driving tasks. (3) The knowledge-driven paradigm induces information of driving scenarios into knowledge-augmented representation space, which can be deduced to generalized knowledge in the scenario semantic space, subsequently inferring the scenarios to guide the drive with the knowledge reflection.}
    \label{fig:three-paradigm}
\end{figure*}

Knowledge is the concretization and generalization of human representation of scenes and events in the real world, representing a summary of experiences and causal reasoning \cite{levesque1986knowledge}. The foundational concepts and significant implications of knowledge-driven approaches can be elucidated through the evolutionary trends in AI. Fig.~\ref{fig:three-paradigm} illustrates different technological paradigms. 
(1) The \textbf{rule-driven paradigm} depends on meticulous logical reasoning or thorough empirical validation using manually crafted rules. These methods aim to encapsulate specific observed phenomena in the real world to facilitate an understanding of driving scenarios from the semantic space. However, handcrafted rules cannot cope with highly complex learning tasks. Moreover, the complexity and diversity of the real world impose evident limitations on these methods, unable to tolerate the fuzziness of continuous space and noisy data \cite{wang2022towards}.
(2) The \textbf{data-driven paradigm} is to establish connectivity-based systems supported by massive data and computational power, capable of emulating the thought processes and world exploration of humans. 
However, the learned representation space processed by data-driven models differs significantly from the scenario semantic space of the human cognitive system, lacking the composability of knowledge and the interpretability of logic \cite{wang2022towards}. Moreover, data-driven models inevitably encounter the data bias or catastrophic forgetting phenomenon \cite{goodfellow2013empirical,kirkpatrick2017overcoming}.
(3) The \textbf{knowledge-driven paradigm} aims to integrate the characteristics of rule-driven and data-driven paradigms, which is a crucial support for propelling significant advancements in the current AI field \cite{wang2022towards, zhang2023toward}. 
Knowledge-driven methods aim to induce information of driving scenarios into knowledge-augmented representation space and deduce to the generalized driving semantic space.
It enables the emulation of human understanding of the real world and the acquisition of learning and reasoning capabilities from experience. Thus, knowledge-driven approaches will be an indispensable pathway for the evolution of the next generation of autonomous driving systems.

Currently, knowledge-driven methods are gradually emerging, with early research endeavors seeking to incorporate knowledge to enhance system performance, particularly in the realm of autonomous driving \cite{tang2021grounded,sur2022domain, bahari2021injecting,lan2022instance, khan2023framework}. However, these studies have not yet been systematically organized and summarized. The knowledge-driven paradigm typically comprises the following key components:

\textbf{Dataset \& Benchmark.} Datasets are digitized perceptions of the real world gathered through various sensors, represented in forms such as images \cite{zablocki2022explainability, huang2022multi}, point clouds \cite{abbasi2022lidar, fei2023self}, etc. The datasets can be endowed with semantic information through manual or automated annotations to construct mechanistic connections between different objects aligning to human cognition \cite{deruyttere2019talk2car, dewangan2023talk2bev, nuscenes, wu2023language, sachdeva2023rank2tell, schick2023toolformer}. The benchmarks established on the datasets serve as evaluation metrics for assessing model performance. It is not only a crucial step in developing data-driven methods but also a prerequisite for constructing large models with general understanding capabilities. However, overemphasizing the inference capabilities of models on datasets may result in the ``overfitting'' dilemma, thereby significantly constraining the models' generalization abilities.

\textbf{Environment.} Environments always serve as cradles for the intelligent agents, providing necessary resource conditions for their survival. The natural world constitutes the only real environment. In contrast to the extended iteration cycles and high trial-and-error costs of the real environment, AI agents can engage in rapid learning and continuous iteration within closed-loop virtual environments. Emerging neural rendering technologies facilitate extensive 3D scene reconstruction at a low cost, creating highly realistic road scenes to robustly support closed-loop environment construction \cite{hu2023pc, wu2023mars, li2023read, guo2023streetsurf, yang2023unisim}. The World Model, designed to model the environment, has the potential to enhance the authentic understanding of driving scenarios, facilitating the progression of autonomous driving from perception to cognition \cite{hu2023gaia, wang2023drivedreamer, min2023uniworld, wang2023driving}. Both neural rendering technologies and world models can facilitate the realization of closed-loop virtual simulations to effectively generate rare corner cases that are difficult to capture in the real world \cite{li2023data, wang2023driving}.

\textbf{Driver Agent.} Knowledge-driven methods shift from passive, data-centric learning to active, cognition-based understanding of the world by systematically applying domain knowledge and reasoning capabilities \cite{muhammad2022vision, fan2022cognitive, li2022deep}. This transformation enables autonomous driving to effectively understand and adapt to unseen driving scenarios \cite{cui2023drivellm, xu2023drivegpt4, mikhailov2023optimizing}. As possessing rich human driving experience and common sense, Large Language Models (LLMs) are commonly employed as foundation models for knowledge-driven autonomous driving nowadays to actively understand, interact, acquire knowledge, and reason from driving scenarios \cite{bubeck2023sparks, schick2023toolformer, jin2023surrealdriver, park2023generative}. Similar to embodied AI's standpoint, true intelligence can only be achieved by curiosity-driven first-person intelligence in the environment \cite{peng2023tong, gildert2023building}. Intelligent agents can continually explore and comprehend their surroundings to support autonomous decision-making and creativity. Analogous to embodied AI, the driving agent should possess the ability to interact with the driving environment, engaging in exploration, understanding, memory, and reflection to achieve genuine intelligence \cite{wen2023dilu, park2023generative}.

\textbf{The objective of this paper is to comprehensively summarize the emerging technological trend involved with knowledge-driven autonomous driving}. We delve into the system framework and core components of knowledge-driven autonomous driving, subsequently analyzing the opportunities and challenges in this field. This paper seeks to provide valuable insights for future research and practical application of autonomous driving, striving to steer its development towards greater safety, reliability, and efficiency.

\else
\section{Introduction}
\label{sec:introduction}
\fi

% \newpage
% \tableofcontents
% \newpage

\section{What is and Why Knowledge-driven Autonomous Driving?}
This section delineates the advantages of knowledge-driven approaches over data-driven methods, illustrated through examples drawn from the evolution of Computer Vision (CV) technologies. Subsequently, we discuss the surge in the development of knowledge-driven techniques driven by generative models like LLMs, and emphasize the significance of data-driven methods in the advancement of autonomous driving.
 
\subsection{Paradigm: Data-driven vs. Knowledge-driven}
\textbf{Limitations of Data-Driven Paradigm.} While existing autonomous driving systems have achieved success in many aspects under the data-driven paradigm, they still struggle to adapt to new driving situations, suffer from overfitting issues caused by data bias, and cannot explain their decisions, ultimately failing to reach a satisfactory level of autonomous driving. The main reason behind these limitations is that data-driven methods emphasize training for specific domains and typically result in systems that excel at the training datasets~\cite{yin2021center, Guo2023scenedm, li2022homogeneous}, but exhibit weak generalization and scalability \cite{zhang2023resimad, pan2017virtual, li2022domain}. This inherent limitation presents a formidable obstacle for autonomous driving systems in coping with the diverse and unpredictable corner cases that frequently arise in real-world driving scenarios \cite{bogdoll2022one, wang2022parallel}.

\textbf{Advantages of Knowledge-driven Paradigm.}
In contrast to traditional data-driven methods, knowledge-driven autonomous driving enables vehicles to have a comprehensive understanding of their surroundings. Essentially, knowledge-driven autonomous driving involves a reasoned, knowledge-based understanding of the real world, enabling it to handle various complex driving scenarios and adapt to ever-changing environments. This understanding involves not only object detection but also semantics understanding  \cite{muhammad2022vision} and context-aware relationship reasoning within the environment \cite{fernandez2019contextual},  to solve complex problems such as multitasking learning and end-to-end learning \cite{ishihara2021multi, casas2021mp3}. Furthermore, the recent emergence of research related to the world model is an advanced form of scenario understanding \cite{mitchell2023ai, lecun2022path, wang2023drivedreamer, hu2023gaia, zhang2023learning}, which is capable of understanding the world and even generating predictions of future world content. The knowledge-driven paradigm can improve system interpretability and trustworthiness, making it easier for human to comprehend the decisions and actions of autonomous driving.

\definecolor{darkspringgreen}{rgb}{0.09, 0.45, 0.27}
\begin{figure*}[tbp]
    \centering
\includegraphics[width=0.85\textwidth]{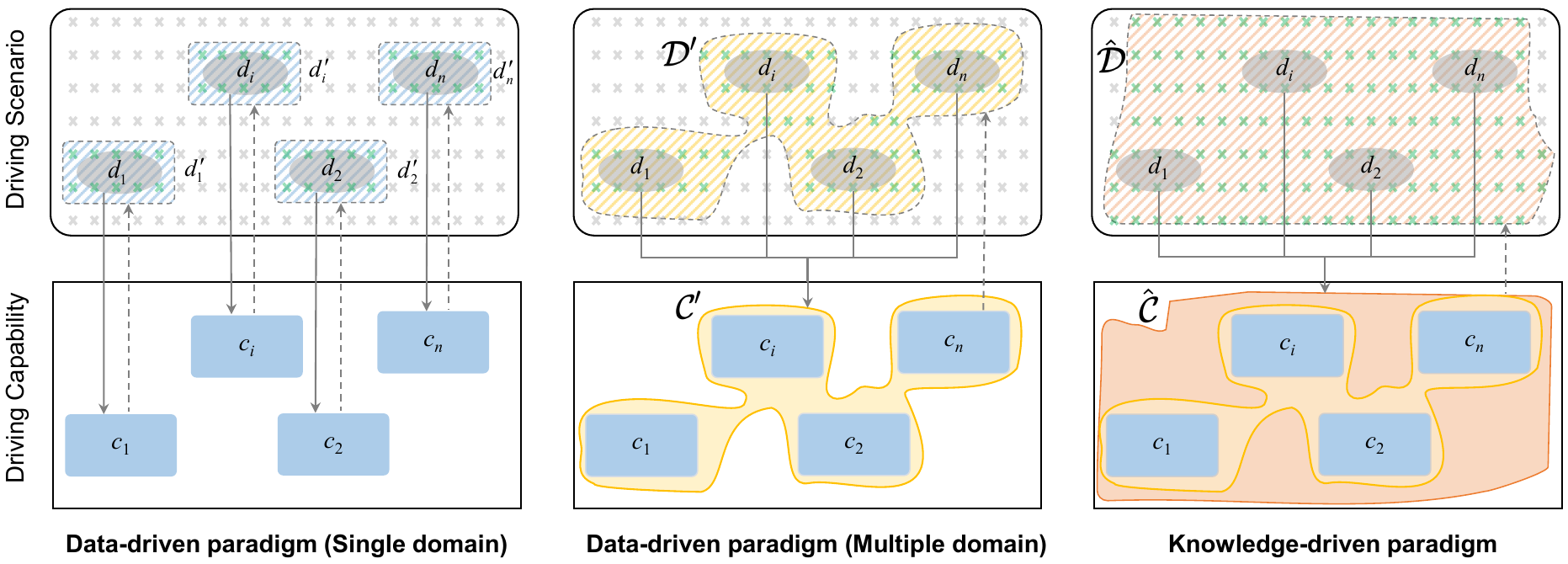}

    \caption{Comparison between the single-domain data-driven paradigm (left), cross-domain data-driven paradigm (center), and the knowledge-driven paradigm (right). The {\color{gray}  gray $\times$} in the driving scenario represents corner cases, while it transitions to \textcolor{darkspringgreen}{green $\times$ }, indicating that the method can handle them respectively. Data-driven approaches focus on collecting domain-specific data $d_i$ and obtaining driving capabilities $c_i$  that are limited to handling only similar or corresponding domains $d_i'$. Even if implementing multiple domains data-driven approaches, it only can learn the driving capability $C'$ for processing the union of datasets $D'$. In contrast, knowledge-driven approaches aim to understand coherent features across domains by incorporating human knowledge or common sense and to establish relationships between features, which achieve a broader range of driving capabilities $\hat{C}$ that far exceed the performances of single-domain data-driven and cross-domain data-driven methods, i.e., $\hat{D}\gg D' > \{d_1, d_2, \ldots, d_n\}$.}
    \label{fig_compare}
    % \vspace{-2ex}
\end{figure*}

The prevailing belief asserts that attaining human-like driving capabilities is pivotal in realizing autonomous driving \cite{schwarting2019social, xia2020human}. Data-driven approaches tend to learn different driving abilities from various driving scenarios, whereas their performances are constrained by the size of the collected dataset \cite{dubois2000knowledge}. Data-driven methods only fit the inputs and outputs of the dataset for a few specific tasks, which makes the acquired capabilities only able to deal with driving scenarios that are closely related to the collected dataset, and cannot generalize and scale to other unseen scenarios. Nevertheless, as the volume of collected data grows, the coverage possibility for new corner cases diminishes, and the marginal effect of capability enhancement becomes increasingly pronounced \cite{o2018scalable, yan2023learning}. In contrast, knowledge-driven methods incorporate human knowledge and common sense into the autonomous driving system, facilitating the establishment of interconnections between different driving domains derived from real-world driving scenarios. Analogous to how a human only needs to have seen an ostrich in a zoo to recognize an ostrich running on a road, the knowledge-driven methods enable understanding and decision reasoning for complex autonomous driving scenarios through generalized scenario understanding capabilities acquired in other domains \cite{kothawade2021auto, fu2023drive}.  Therefore, this approach is anticipated to bridge the gap between various driving domains, ultimately resulting in more generalized driving capabilities. Remarkably, the capabilities derived from this paradigm demonstrate the capacity to drive in broader domains compared to those obtained through data-driven approaches. This concept is further elucidated in Fig. \ref{fig_compare}: although data-driven approaches can acquire driving capability by extracting features from datasets, both single-domain learning and multi-domain learning are abstractions in high-dimensional spaces, $c_i$ or $\cal{C}'$, with limited generalization capabilities. While knowledge-driven approaches can compress the driving capability space $\hat{\cal{C}}$ into a low-dimensional manifold space \cite{saul2003think} by summarizing the experiences from multi-domain data to construct foundational models with general comprehension capabilities. The driving scenario corresponding to this space not only includes the data collected during training, but also covers a lot of unseen data, including a large number of corner cases.

\subsection{Exploring the Knowledge-driven Trend in CV Tasks}

In recent years, the traditional computer vision community has witnessed a significant transformation, shifting from the perceptive paradigm to the cognitive paradigm. In the earlier phase, data-driven methods predominantly focused on task completion without a profound understanding of the underlying semantics. This resulted in models that were effective at discrimination tasks but lacked true comprehension of the data, like Image Classification has historically been a cornerstone of CV. Traditional data-driven methods, such as Convolutional Neural Networks (CNNs), focused on training models to recognize and categorize images. These methods excelled at specific tasks including handwritten digits classification~\cite{deng2012mnist} and pioneering classification task research on ImageNet~\cite{deng2009imagenet}. Data-driven approaches for 2D Object Detection~\cite{zou2023object} aimed to locate and classify objects within images. Methods like Faster R-CNN~\cite{ren2015faster} and YOLO~\cite{redmon2016you} were widely adopted for this purpose. However, these methods primarily emphasized task performance without deep semantic understanding. And Semantic/Instance Segmentation involves identifying object boundaries and their categories. Techniques like U-Net~\cite{ronneberger2015u} and Mask R-CNN~\cite{he2017mask} are representative of data-driven approaches that excelled at segmentation tasks but did not emphasize semantic comprehension.

In contrast, knowledge-driven approaches aim to empower CV tasks with a deeper understanding of semantics and recognition. To address the issue that traditional methods fail to genuinely understand data, some research has shifted towards training generative models or combining multimodal data to learn more robust data representations. For instance, Image Captioning~\cite{you2016image} attempts to make models comprehend the content of images and generate descriptive text, thereby demonstrating the model's true understanding of the image content. The Visual Question Answering (VQA)~\cite{antol2015vqa} verifies the model's reasoning ability by constructing complex question-answer pairs related to image content. There are even datasets like Visual Genome~\cite{krishna2017visual} that can perform multiple complex tasks such as object detection, image description, and object relationship inference simultaneously. Moreover, with the increase in computational power, research in this domain has expanded from images to videos. Until now, research in the field of CV remains dynamic. The emergence of Generative Adversarial Networks (GAN)~\cite{goodfellow2014generative,pan2023drag} and Variational Autoencoders (VAE)~\cite{kingma2013auto} validates the potential of generative models, while the Diffusion Model~\cite{ho2020denoising,dhariwal2021diffusion,rombach2022high} has elevated cross-modal understanding to a new level.

\subsection{LLM: A Milestone for Knowledge-driven Approaches}
Recently, LLMs have achieved remarkable performance. These models have achieved remarkable performance by leveraging extensive training on massive text datasets, showcasing powerful text generation and comprehension capabilities. LLMs have demonstrated their competence in understanding natural language and tackling diverse complex tasks~\cite{zhao2023survey}, emerging as a milestone in the development of knowledge-driven methods. 
Some notable examples of LLMs include GPT-3~\cite{floridi2020gpt}, PaLM~\cite{chowdhery2022palm}, LLaMA~\cite{touvron2023llama}, and GPT-4~\cite{openai2023gpt4}. Notably, the emergent capability in LLMs is one of their most distinguishing features compared to smaller language models. Specifically, capabilities such as contextual learning~\cite{brown2020language}, instruction following~\cite{ouyang2022training,wei2021finetuned}, and chain of thought reasoning~\cite{wei2023chainofthought} are three typical emergent abilities in LLMs. Specifically, ChatGPT~\cite{chatgpt} and GPT-4~\cite{openai2023gpt4} represent significant advancements in LLM capabilities, especially in natural language understanding and generation. It's worth noting that LLMs are seen as equipped with human-like intelligence and common sense to hold the potential to bring us closer to the field of Artificial General Intelligence (AGI)~\cite{zhao2023survey, zhu2023minigpt}. Remarkable breakthroughs in LLMs underscore the critical importance of high-quality data. These models exhibit robust reasoning capabilities and also possess emergent capacity, which lays a solid foundation for the development of knowledge-driven autonomous driving.

\subsection{Significance of Knowledge-driven Methods to Autonomous Driving}

Data is critical to the development of autonomous driving technology, which relies on massive amounts of data to optimize algorithmic models to be able to recognize and understand the road environment to make the right decisions and actions \cite{yurtsever2020survey, chen2022milestonessurvey}. For example, the huge amount of data and driving scenarios accumulated by Tesla is an important reason for being able to stay ahead of the curve in autonomous driving algorithms. As the autonomous driving task is evolving from a single perception task to an integrated multi-task of perception and decision-making \cite{chen2023end}, the diversity and richness of autonomous driving data modalities are becoming critical. However, models trained solely on large amounts of collected data can only be third-person intelligence \cite{peng2023tong, gildert2023building}, which refers to an AI system that observes, analyzes, and evaluates human behaviors and performances from a bystander's perspective. However, the ultimate form of autonomous driving will be the realization of a generalized AI for the driving domain \cite{liu2020overview, dou2023towards}, which makes the shift from the data-driven paradigm to the knowledge-driven paradigm an inevitable requirement for the evolution of autonomous driving.

\begin{figure*}[tbp]
    \centering
    \includegraphics[width=0.85\textwidth]{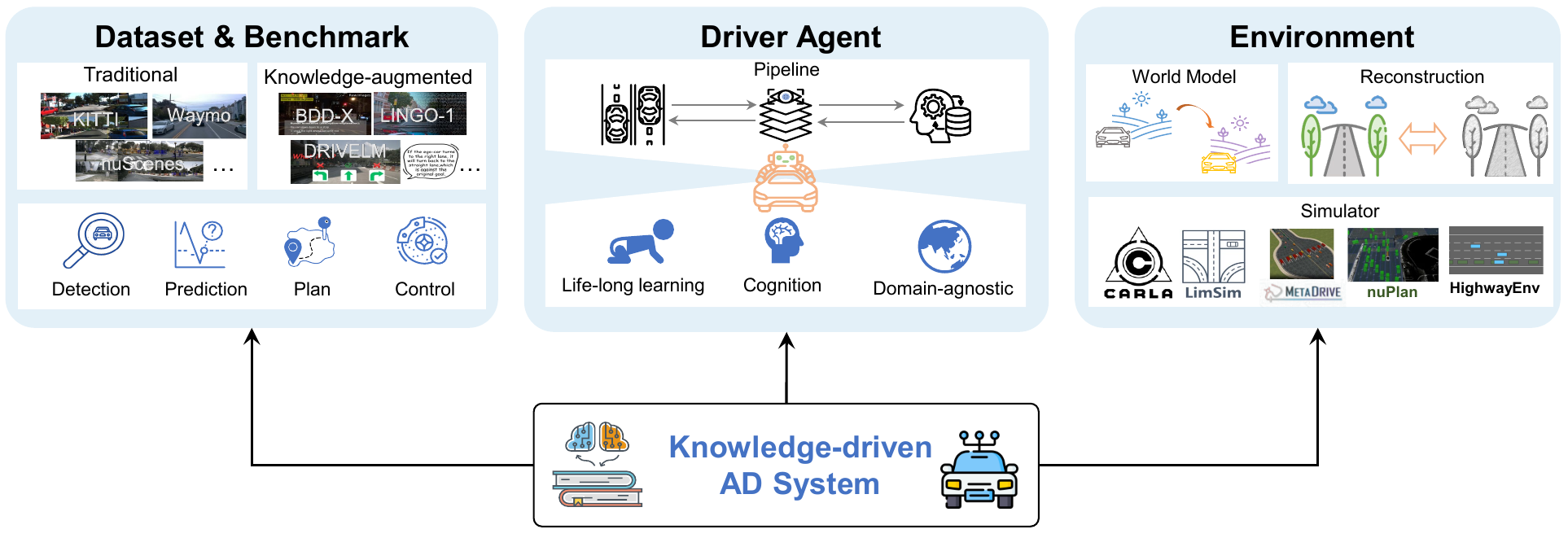}
    \caption{Key components in knowledge-driven autonomous driving.} 
    \label{fig_knowledgead}
\end{figure*}

The knowledge-driven paradigm does not completely detach from the original data-driven approaches but adds the design of knowledge or common sense based on the data-driven approaches, such as common sense judgment, empirical induction, logical reasoning, etc. Knowledge-driven methods rely on AI agents to explore the environment and acquire general knowledge, as opposed to the implementation of predefined human rules or the portrayal of abstract characteristics from collected data \cite{mitchell2023ai, lecun2022path}. Specifically, the iterative updating of the knowledge-driven approach requires the continuous summarization of data from the Agent's interaction with the environment to form new specialized domain knowledge to enhance the specialized capabilities \cite{zhang2023toward, xi2023rise}. Recent advancements in autonomous driving reflect this shift from purely data-driven methodologies to those that derivation from knowledge-driven.

\textbf{Transformation of Perception Module.} Previous autonomous driving perception modules usually perform open-loop fitting on a dataset to recognize and localize semantic information in the scene, including 3D object detection~\cite{yin2021center,li2022bevformer,liu2023bevfusion}, lane detection~\cite{hou2019learning,chen2022persformer}, semantic segmentation~\cite{kong2023robo3d,liu2023uniseg}, etc. The inputs to the perception module are usually pictures captured by cameras and point cloud information collected by LiDAR. Correspondingly, there are camera-only~\cite{li2022bevformer,huang2021bevdet}, LiDAR-only~\cite{lang2019pointpillars,yin2021center,deng2021voxel}, and LiDAR-camera fusion~\cite{bai2022transfusion,liu2023bevfusion,li2023logonet} schemes for perception methods. Recently, many scholars have realized that a full understanding of the environment requires a shift from perception to cognition. Since the setup of in-vehicle sensors is often a comprehensive coverage of multiple types and perspectives, the multimodal data collected needs to be semantically aligned in a high-dimensional space to realize a true understanding of the driving scene~\cite{Wayve-LINGO-1,ma2023dolphins}.

\textbf{Knowledge-embedded Decision-making and Planning.} Early automated driving decision planning was usually done by building explicit mathematical models for fitting driving data, including the classical car-following and lane-changing models \cite{gazis1961nonlinear, treiber2000congested, kesting2007general}. To improve the applicability of the models in different scenarios, these explicit mathematical models need to be continuously improved based on expert knowledge and increase the complexity of the models. However, the diversity of real-world scenarios makes such improvements increasingly challenging. As a result, researchers often resort to manually designed state machines to address as many corner cases encountered in real-vehicle testing as possible \cite{hulnhagen2010maneuver, bae2020finite,bolte2019towards}. 
Contrastingly, another category of modeling concepts aims to harness the exploratory capabilities of heuristic search methods and the approximation power of deep learning, with the goal of surmounting challenges associated with manual design. Despite these efforts, these approaches continue to encounter difficulties in complex scenarios. Heuristic search methods heavily rely on human-designed heuristic functions, and the dimension explosion also poses a challenge in achieving approximate optimal solutions within finite time \cite{ma2014fast, wen2023trafficmcts}. Reinforcement learning methods require closed-loop training in simulation engines or even real environments at a high cost and expense, and the convergence conditions of the model often depend on the reasonableness of the manually designed reward function \cite{guo2021hierarchical, rupprecht2022survey}. Although it is possible to obtain the reward function from the data by inverse reinforcement learning methods \cite{arora2021survey}, it also means that the model is less capable of generalizing to different environments. Incorporating human knowledge to support autonomous driving also presents a significant challenge for decision planning. Compared with the insurmountable limitations of other decision planning models, including social force-based models \cite{helbing1995social, yang2018social}, risk field-based models \cite{wang2015driving, wang2016driving}, etc., the powerful knowledge utilization and reasoning capabilities recently demonstrated by LLMs are more suitable for understanding, reasoning, and decision making for autonomous driving \cite{liu2023can, fu2023drive, cui2023survey}.

\textbf{The Trend towards Modular Convergence.} The end-to-end technology route was also the plain idea of early research in autonomous driving. For example, CMU's Navlab implemented an autonomous driving system based on an end-to-end model as early as the 1980s \cite{pomerleau1988alvinn}, which used visual sensor data as inputs and directly outputted steering wheel angle, brake pedal strength, and other in-line signals to control the vehicle. However, this was limited by the uncertainty brought by the arithmetic conditions and black-box system at that time. With the diversified and uneven development of autonomous driving perception, planning, control, and other technologies, emerging autonomous driving companies represented by Tesla and Waymo have gradually constructed a modular-based autonomous driving pipeline \cite{schwarting2018planning, ma2020artificial}, which has become prevalent autonomous driving solutions. Subsequently, perception, planning, and decision-making have shown a trend of convergence, including the integration of prediction and decision-making, and even end-to-end autonomous driving \cite{daudelin2018integrated, casas2021mp3, sadat2020perceive, hu2023planning}. Researchers generally realize that autonomous driving is oriented to the ultimate goal of vehicle performance such as safety and efficiency \cite{vahidi2018energy, wang2020cooperative}. End-to-end autonomous driving can avoid overall performance degradation due to heterogeneous optimization directions and cascading information transfer errors \cite{chen2023end}. 

From a knowledge-driven perspective, perception, prediction, planning, and control have a sequential causal relationship, which is easily evidenced in common driving scenarios. For instance, a cyclist turning back in a non-motorized lane could indicate an intention to make a turn, and a vehicle activating its turn signal while proceeding straight may signify an upcoming lane change within a few seconds.
The separate perception modules, which merely convey bounding boxes to the prediction and decision-making modules, present challenges in ensuring the subsequent modules' performance effectiveness. In contrast, the end-to-end frameworks based on module fusion can efficiently extract and convey features closely associated with the driving task. However, existing end-to-end frameworks still represent only a high level of abstraction of knowledge and are unable to articulate the utilization of driving knowledge manifested in the model output \cite{hu2023planning, chib2023recent}. Therefore, the textualized explanation of scene understanding and logical reasoning provided by the LLMs is anticipated to enhance the credibility and robustness of the existing end-to-end framework.

In summary, the knowledge-driven paradigm stands at the forefront of recent advancements in autonomous driving technology. When equipped with high-quality data and a suitable environmental platform, the pivotal question becomes the design of effective knowledge-driven modeling solutions. This entails integrating human driving experience and common sense into the system, developing knowledge models endowed with the capability to reason and solve intricate driving challenges. Knowledge-driven modeling approaches empower autonomous driving systems to adeptly navigate evolving traffic and road scenarios, thereby enhancing system performance, interpretability, and safety.
%In the next sections, we will introduce the knowledge-driven system framework synthesized in this paper comprises the following key components as shown in ~\ref{fig_knowledgead}, and we expand on the datasets, benchmarks, how to build high-quality environments, and how to obtain knowledge-driven driver agents related to knowledge-driven autonomous driving.
In the following sections, we will introduce the knowledge-driven system framework synthesized with key components, as illustrated in Fig. ~\ref{fig_knowledgead}. This includes the development of datasets and benchmarks, how to construct high-quality environments, and how to acquire knowledge-driven driver agents for autonomous driving.

\section{Datasets \& Benchmarks}

The safety and reliability of autonomous driving systems have always been crucial evaluation factors. For the evaluation of knowledge-driven autonomous driving, researchers develop and assess these systems using appropriate datasets, benchmarks, and metrics. Traditional data-driven autonomous driving datasets~\cite{kitti-dataset,cityscapes,bddv,hdd,wod,nuscenes,panonus} provide mappings from sensor data to perception, prediction, and planning labels. Accompanied by the emergence of knowledge-driven autonomous driving, various groups of researchers augment preexisting~\cite{refer-kitti,cityscapes-ref,bddx,had,nuscenes-qa,drivelm,deruyttere2019talk2car} or recently acquired datasets~\cite{dreye,dada2000,HDBD-inco,drama,sachdeva2023rank2tell} with different types of knowledge, mainly in the modality of natural language and gaze heatmap. By incorporating external knowledge, the intelligence level of autonomous driving models gradually evolves from perception level to cognition level, ensuring stronger reliability and interpretability. This section first introduces traditional autonomous driving datasets and then delves into existing knowledge-augmented autonomous driving datasets and corresponding benchmarking tasks. Finally, this section presents commonly used tasks and evaluation metrics in knowledge-oriented autonomous driving benchmarks.

\begin{figure*}[tbp]
    \centering
    \includegraphics[width=0.62\textwidth]{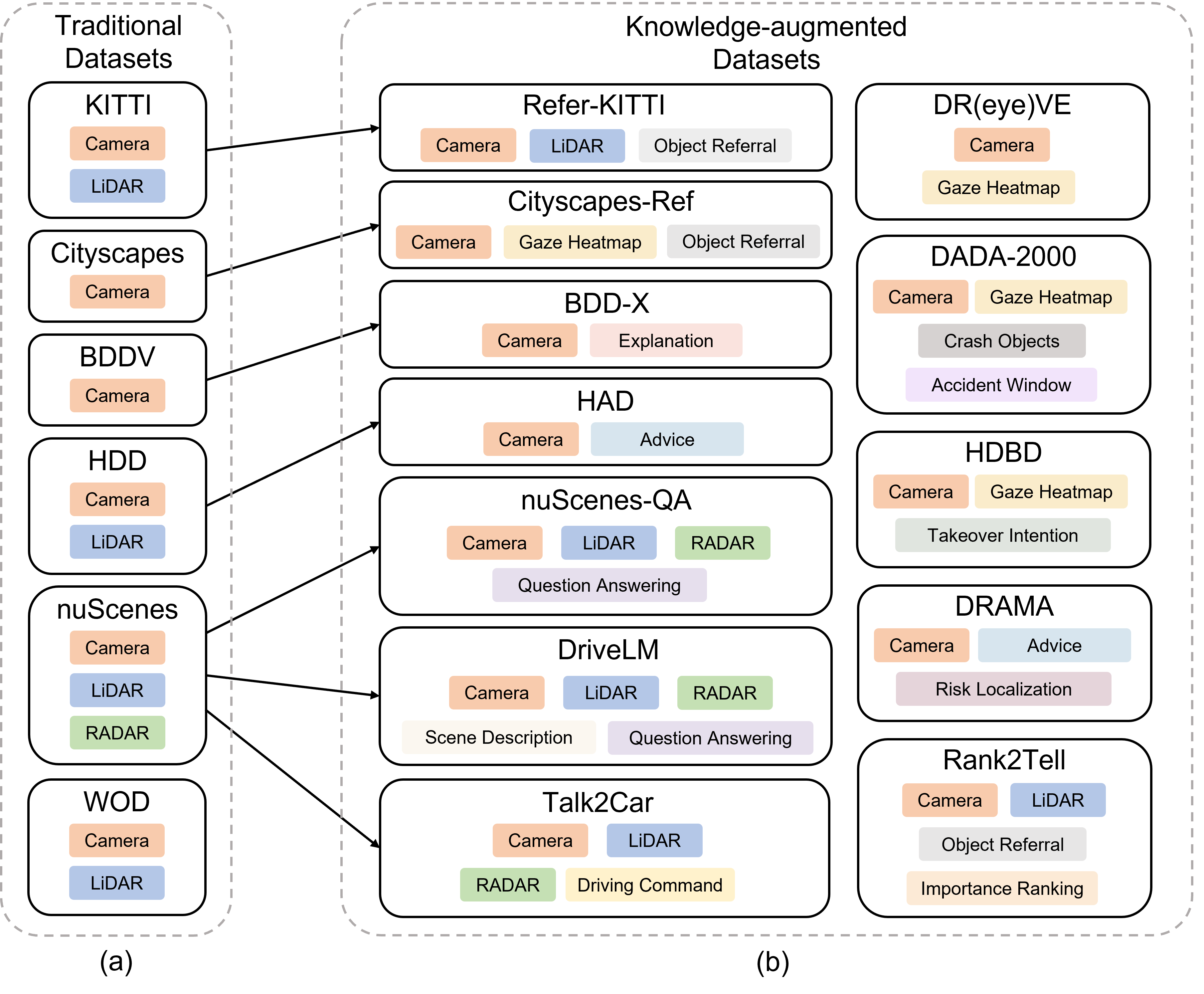}
    \caption{(a) Traditional and (b) knowledge-augmented autonomous driving datasets. The arrow indicates that the knowledge-augmented datasets are derived from the corresponding source dataset through secondary annotation.}
    \label{fig_data}
\end{figure*}

\subsection{Traditional Datasets}

This section provides a detailed introduction to traditional autonomous driving datasets, which are also visualized in Fig.~\ref{fig_data}(a).

\textbf{KITTI dataset}~\cite{kitti-dataset} is a collection of sensor data recorded in and around Karlsruhe, Germany, with the main purpose of advancing computer vision and robotic algorithms for autonomous driving. It includes camera images, laser scans, high-precision GPS measurements, and IMU accelerations. The dataset provides precise instructions for accessing the data and insights into sensor limitations and common challenges. The sensor setup consists of grayscale and color cameras, a 3D laser scanner, and an inertial and GPS navigation system. The dataset is divided into categories such as ``Road'', ``City'', ``Residential'', ``Campus'', and ``Person'', and includes raw data, object annotations in the form of 3D bounding boxes, tracklets, and calibration data.

\textbf{Cityscapes dataset}~\cite{cityscapes} is a large-scale benchmark suite for semantic urban scene understanding. It consists of stereo video sequences captured from a moving vehicle in 50 different cities, primarily in Germany. The dataset includes 5,000 images with high-quality pixel-level annotations and an additional 20,000 images with coarse annotations. The data recording and annotation methodology were designed to capture the variability of outdoor street scenes. The dataset provides both fine and coarse pixel-level annotations for 30 visual classes, including instance-level labels for humans and vehicles. The annotations were carefully quality controlled, and the dataset includes vehicle odometry, outside temperature, and GPS tracks. The Cityscapes dataset surpasses previous attempts in terms of size, annotation quality, scene variability, and complexity.

\textbf{Berkeley DeepDrive Video dataset (BDDV)}~\cite{bddv} is a large and diverse dataset consisting of real driving videos and GPS/IMU data. It covers various driving scenarios such as cities, highways, towns, and rural areas in major US cities. Compared to earlier datasets like KITTI and Cityscapes, BDDV stands out in terms of its scale, providing over 10,000 hours of driving videos. Additionally, BDDV includes smartphone sensor data such as GPS, IMU, gyroscope, and magnetometer readings, which can be used to analyze vehicle trajectory and dynamics. The dataset aims to capture the diversity of driving scenes, car makes and models, and driving behaviors. This makes BDDV suitable for learning a generic driving model.

\textbf{Honda Research Institute Driving dataset (HDD)}~\cite{hdd} is a collection of sensor data recorded from an instrumented vehicle in the San Francisco Bay Area. The dataset includes video from three cameras, 3D LiDAR data, GPS signals, and signals from the vehicle's CAN bus. The data collection aimed to capture diverse traffic scenes and driver behaviors. The dataset consists of 104 hours of video, with annotations based on a 4-layer representation of driver behavior. The annotation methodology incorporates both objective criteria and subjective judgment. The dataset provides insights into driver behavior, including goal-oriented actions, stimulus-driven actions, causes, and attention. The dataset is around 150GB in size including 137 sessions with an average duration of 45 minutes.

\textbf{nuScenes dataset}~\cite{nuscenes} is a collection of driving data from Boston and Singapore, featuring diverse locations, weather conditions, and driving scenarios. The dataset includes 84 logs with 15 hours of driving data, captured using Renault Zoe electric cars equipped with various sensors. The data is carefully synchronized, and localization is achieved through a robust LiDAR-based method. Highly accurate human-annotated semantic maps and baseline routes are provided. The dataset contains 1000 interesting scenes manually selected, covering high traffic density, rare events, and challenging situations. Expert annotators provide detailed annotations for 23 object classes, including pedestrians and vehicles. The dataset encourages research on long-tail problems and offers high-frequency sensor frames.
% making it unique in capture frequency.

\textbf{Waymo Open dataset (WOD)}~\cite{wod} provides sensor data collected using five LiDAR sensors and five high-resolution pinhole cameras. The LiDAR data includes the first two returns of each laser pulse, while the camera images are captured using rolling shutter scanning. The dataset offers ground truth annotations for both LiDAR and camera data, including 3D bounding boxes for objects in LiDAR data and 2D bounding boxes for objects in camera images. Multiple coordinate systems are used, such as global, vehicle, sensor, and image frames. The dataset covers suburban and urban areas, with approximately 12 million labeled 3D LiDAR objects and 12 million labeled 2D image objects.

\begin{table*}[t]
\renewcommand\arraystretch{1.5}
\centering
\footnotesize
\caption{Key attributes of existing knowledge-augmented datasets. C, L, and R stand for Camera, LiDAR, and Radar respectively.}
\label{table_data}
\begin{tabular}{l|c|p{4cm}|p{4cm}|p{4cm}}
\toprule[1.5pt]
\multicolumn{1}{c|}{\centering\textbf{Dataset}} & 
\multicolumn{1}{c|}{\centering\textbf{Sensors}} & \multicolumn{1}{c|}{\centering\textbf{Knowledge Form}} & \multicolumn{1}{c|}{\centering\textbf{Tasks}} & \multicolumn{1}{c}{\centering\textbf{Metrics}} \\

\Xhline{1pt}
BDD-X~\cite{bddx}& C & Explanation & Vehicle Control, Explanation Generation, Scene Captioning & MAE, MDC, BLEU-4, METEOR, CIDEr-D \\
\Xhline{1pt}
Cityscapes-Ref~\cite{cityscapes-ref}& C & Object Referral, Gaze Heatmap & Object Referring & Acc@1 \\
\Xhline{1pt}
DR(eye)VE~\cite{dreye}& C & Gaze Heatmap & Gaze Prediction & CC, KLD, IG \\
\Xhline{1pt}
HAD~\cite{had}& C & Advice & Vehicle Control & MAE, MDC \\
\Xhline{1pt}
Talk2Car~\cite{deruyttere2019talk2car}& C+L+R & Object Referral & Object Referring & IoU@0.5 \\
\Xhline{1pt}
DADA-2000~\cite{dada2000}& C & Gaze Heatmap, Crash Objects, Accident Window & Gaze Prediction & CC, KLD, NSS, SIM \\
\Xhline{1pt}
HDBD~\cite{HDBD-inco}& C & Gaze Heatmap, Takeover Intention & Driver Takeover Detection & AUC \\
\Xhline{1pt}
Refer-KITTI~\cite{refer-kitti}& C+L & Object Referral & Object Referring, Object Tracking & HOTA \\
\Xhline{1pt}
DRAMA~\cite{drama}& C & Advice, Risk Localization & Motion Planning & L2 Error, Collision Rate \\
\Xhline{1pt}
Rank2Tell~\cite{sachdeva2023rank2tell}& C+L & Object Referral, Importance Ranking & Importance Estimation, Scene Captioning & F1 Score, Accuracy, BLEU-4, METEOR, ROUGE, CIDER \\
\Xhline{1pt}
DriveLM~\cite{drivelm}& C+L+R & Scene Captioning, Question Answering & Scene Captioning, Question Answering & ADE, FDE,  Accuracy,  Collision Rate, SPICE, GPT-Score \\
\Xhline{1pt}
NuScenes-QA~\cite{nuscenes-qa}& C+L+R & Question Answering & Question Answering & Exist, Count, Object, Status, Comparison, Acc \\
\bottomrule[1.5pt]

\end{tabular}
\vspace{10pt}
\end{table*}

\subsection{Knowledge-augmented Datasets}

This section provides a detailed introduction to knowledge-augmented autonomous driving datasets, which are also visualized in Fig.~\ref{fig_data}(b). Additionally, Table~\ref{table_data} presents key attributes of existing knowledge-augmented datasets.

\textbf{Berkeley DeepDrive eXplanation (BDD-X) dataset}~\cite{bddx} contains over 77 hours of driving videos with accompanying textual justifications for driving actions. It includes diverse driving conditions and activities, such as lane changes and turns, annotated by human annotators familiar with US driving rules. It consists of a training set, a validation set, and a test set with a total of 6,984 videos. BDD-X dataset aims to improve the trust and user-friendliness of self-driving cars by providing explanations for their decisions. To fulfill this goal, the dataset utilizes three benchmarking tasks, namely vehicle control, explanation generation, and scene captioning.

\textbf{Cityscapes-Ref dataset}~\cite{cityscapes-ref} focuses on object referring in videos, incorporating language descriptions and human gaze. It includes 5,000 stereo video sequences from the Cityscapes dataset, with annotations for object descriptions, bounding boxes, and gaze recordings. The dataset aims to address the limitations of previous datasets by providing temporal, spatial context, and gaze information. Cityscapes-Ref dataset employs the task of object referring for benchmarking.

\textbf{DR(eye)VE dataset}~\cite{dreye,dreyeproj} consists of 555,000 frames from 74 sequences, captured during a driving experiment with eight drivers in various contexts and weather conditions. The dataset includes eye-tracking data from SMI ETG glasses and car-centric views from a roof-mounted camera. The dataset enables the analysis of driver behavior and attention in real-life driving scenarios. Fixation maps are computed using a temporal sliding window, and attention drifts are labeled for evaluation purposes. Multiple baselines are tested on DR(eye)VE dataset for the task of gaze prediction.

\textbf{Honda Research Institute-Advice dataset.} Honda Research Institute-Advice Dataset (HAD)~\cite{had} consists of 5,675 driving video clips with human-annotated textual advice. The videos are collected from the HDD dataset~\cite{hdd} and include various driving activities in urban settings. Annotators describe the driver's actions and provide attention descriptions from a driving instructor's perspective. The dataset contains a total of 25,549 action descriptions and 20,080 attention descriptions. The advice covers topics such as speed, driving maneuvers, traffic conditions, and road elements. Multiple baseline methods are evaluated on HAD dataset for the task of vehicle control.

\textbf{Talk2Car dataset}~\cite{deruyttere2019talk2car} is built upon the nuScenes dataset and includes 850 videos with written commands for autonomous driving. The dataset covers different cities, weather conditions, and times of day. Each video has annotations for six cameras, LIDAR, GPS, IMU, RADAR, and 3D bounding boxes for 23 object classes. The dataset contains 11,959 commands, with an average of 11.01 words per command. The dataset provides a wide distribution of commands, object distances, and object categories. Talk2Car dataset employs the task of object referring for benchmarking.

\textbf{DADA-2000 dataset}~\cite{dada2000,dada} is a collection of accident videos obtained from various video websites. It consists of 658,476 frames from 2000 videos, covering a duration of 6.1 hours. The dataset includes diverse accident categories and provides annotations for spatial crash objects, temporal accident windows, and attention maps. It offers a comprehensive representation of accident situations in driving scenes and is more complex compared to previous datasets for driving accident analysis. This dataset utilizes gaze prediction as the primary benchmarking task.

\textbf{HRI Driver Behavior dataset (HDBD)}~\cite{HDBD-inco} contains driver behavior data collected from simulator and real scene videos. The dataset includes behavioral and physiological signals from 28 participants, along with environmental and vehicle sensory information. The data was collected using eye-tracking devices, physiological sensors, and vehicle/driving simulator sensory data. The dataset includes human-AV interaction data from 32 participants, focusing on monitoring L2 automated driving through intersections. The dataset provides information on takeover intention, HMI transparency levels, maneuvers, weather conditions, and synchronized signals for analysis.  Authors evaluate multiple baseline methods on HDBD dataset for driver takeover detection task.

\textbf{Refer-KITTI dataset}~\cite{refer-kitti} is a dataset constructed based on the public KITTI dataset~\cite{kitti-dataset}, aimed at referring understanding. It utilizes instance-level box annotations from KITTI and a labeling tool to efficiently annotate referent objects across frames. The dataset features diverse scenes and provides descriptive statistics on object numbers and temporal dynamics. Refer-KITTI includes 818 expressions and is split into 15 training videos and 3 testing videos, offering flexibility and temporal challenges for referent object association. This dataset utilizes object referring and tracking as the primary benchmarking task.

\textbf{DRAMA dataset}~\cite{drama} is designed for evaluating visual reasoning capabilities in driving scenarios. It consists of 17,785 interactive driving scenarios recorded from urban roads in Tokyo. The dataset includes synchronized videos, CAN signals, and IMU information. Annotations are provided through object-level and video-level questions and answers, focusing on identifying important objects and generating associated attributes and captions. The dataset statistics highlight the distribution of labels, object types, visual attributes, and reasoning descriptions. DRAMA dataset utilizes motion planning as the primary benchmarking task.

\textbf{Rank2Tell dataset}~\cite{sachdeva2023rank2tell} consists of 116 video clips captured at intersections using multiple cameras, LiDAR sensors, and GPS in diverse traffic scenes. The dataset focuses on identifying and ranking important agents that can influence the ego vehicle's driving. Annotations are provided by five annotators, considering agent identification, localization, ranking, and captioning. The dataset emphasizes explainability by providing captions that explain why agents are deemed significant. The dataset enables the evaluation of agent importance perception and caption diversity in traffic scenes. This dataset employs two benchmarking tasks, namely importance estimation and scene captioning.

\textbf{DriveLM dataset}~\cite{drivelm} is an autonomous driving dataset that connects LLMs and autonomous driving systems. It incorporates linguistic information and reasoning abilities to facilitate perception, prediction, and planning (P3) in autonomous driving. The dataset includes frame-based QA pairs connected in a graph-style structure, covering perception, prediction, and planning tasks. It is based on the nuScenes dataset and aims to enhance the reasoning and decision-making capabilities of autonomous driving systems. Scene captioning and question answering tasks are incorporated for benchmarking.

\textbf{nuScenes-QA dataset}~\cite{nuscenes-qa} is constructed for 3D question answering in driving scenarios. It combines scene graphs generated from 3D annotations with manually designed question templates to generate question-answer pairs. The dataset contains 459,941 pairs based on 34,149 visual scenes, with a wide range of question types and lengths. It is the largest 3D-related question answering dataset, providing balanced distributions of questions and answers. The dataset poses challenges for models due to its complexity and diverse visual semantics.

\subsection{Benchmarking Tasks and Evaluation Metrics}

This section offers an in-depth overview of various benchmarking tasks and associated evaluation metrics specific to knowledge-driven autonomous driving.

\textbf{Motion Prediction and Planning} involves forecasting the trajectories of various traffic participants (vehicles, pedestrians, etc.), and planning the future movements of an ego vehicle in both open-loop and closed-loop manners. Key metrics for motion prediction include Average Displacement Error (ADE), Final Displacement Error (FDE), Miss Rate, Overlap Rate, Average Heading Error (AHE), and Mean Average Precision (mAP). ADE assesses the average displacement error of the closest prediction to the ground truth trajectory, while FDE evaluates the displacement error at a specific future time step. Miss Rate is determined by whether the model's predictions for traffic participants fall within certain thresholds of the ground truth trajectory. Overlap Rate examines the incidence of predicted trajectories overlapping with other objects in the scenario. AHE is defined as the average of the heading angle differences between the predicted trajectory and the ground truth. mAP provides a comprehensive evaluation by categorizing trajectories and measuring the precision and recall of the predictions against the ground truth.
For the task of open-loop planning, metrics are similar to those of motion prediction, as they involve predicting the ego vehicle's future trajectory. In contrast, closed-loop ego vehicle planning tasks entail following the output trajectory from the method and continuously interacting with traffic participants in the dynamic scene. Key metrics for closed-loop planning tasks typically include No at-fault Collisions, Drivable Area Compliance, Speed Limit Compliance, Comfort, and Time to Collision (TTC) within bounds. These metrics ensure that the ego vehicle's trajectory avoids collisions with other vehicles, drives within the mapped drivable area, and obeys speed limits at all times. Comfort is measured by evaluating the minimum and maximum longitudinal and lateral accelerations of the ego vehicle's driven trajectory.

\textbf{Scene Captioning and Explanation Generation}. 
Given a stream or a frame of sensory data, e.g. camera and/or LiDAR data, these two tasks require the captioning model and explaining model to generate description and reasoning texts. To evaluate the performance of captioning and explaining models, metrics including BLEU~\cite{papineni2002bleu}, METEOR~\cite{lavie2007meteor}, ROUGE~\cite{lin2004rouge}, CIDEr~\cite{vedantam2015cider}, CIDEr-D~\cite{vedantam2015cider}, SPICE~\cite{anderson2016spice} are adopted, whose details are discussed below.
\textit{BLEU}~\cite{papineni2002bleu} is an automatic evaluation metric that measures the similarity between a machine-generated translation and reference translations based on n-gram precision. It calculates the precision of n-grams up to a 4-gram level by counting matching n-grams. The modified precisions for each n-gram length are combined using a weighted geometric mean to compute the BLEU score, which ranges from 0 to 1.
\textit{METEOR}~\cite{lavie2007meteor} is an automatic evaluation metric used to assess the quality of machine-generated translations or text generation systems. It captures overall quality, fluency, and adequacy. METEOR calculates the number of matching unigrams between the machine-generated translation and reference translations, considering exact word matches, stemming, and synonymy. Precision, recall, alignment, and ordering scores are combined using a weighted harmonic mean to obtain the final METEOR score.
\textit{ROUGE}~\cite{lin2004rouge} is an automatic evaluation metric commonly used in NLP to assess text summarization systems. It quantifies the overlap between the generated summary and reference summaries. ROUGE involves preprocessing, n-gram matching, calculation of recall and precision, and computation of the F-measure as the harmonic mean of recall and precision. ROUGE scores are typically computed for multiple n-gram lengths and aggregated to obtain an overall score.
\textit{Consensus-based Image Description Evaluation (CIDEr)}~\cite{vedantam2015cider} is an evaluation metric used in the field of computer vision and image captioning to assess the quality of automatically generated image captions. It aims to capture both the relevance and diversity of the generated captions. CIDEr measures the consensus between the generated captions and the human-generated reference captions. The CIDEr metric computes the similarity between the generated captions and the reference captions based on n-gram matching and term frequency-inverse document frequency (TF-IDF) weighting.
\textit{CIDEr with Diversity (CIDEr-D)}~\cite{anderson2016spice} is an extension of the CIDEr metric that incorporates diversity into the evaluation. It encourages the generation of diverse and informative captions by penalizing captions that are similar to each other. CIDEr-D achieves this by adding a diversity term to the original CIDEr score, which measures the uniqueness of the generated captions.
\textit{SPICE (Semantic Propositional Image Caption Evaluation)} ~\cite{anderson2016spice}, serves as a principled metric for the automated evaluation of image captions. It specifically assesses the similarity in semantic propositional content between predicted captions and ground truth references. It quantifies the structural resemblance between the predicted and reference texts while abstracting away from the consideration of semantic meanings.

\textbf{Object Referring} involves referring to specific objects within images or scenes using natural language descriptions. In object referring, a typical scenario involves an image or a scene accompanied by a textual description that refers to a particular object or region of interest within that visual input. The goal is to develop models that can comprehend the textual description and effectively map it to the corresponding object or region in the image. Commonly used metrics include Acc@1 and IoU@0.5.
\textit{Acc@1} metric is a commonly used evaluation measure to assess the performance of models in accurately localizing or identifying referred objects. Formally, let $N$ denote the total number of object referral instances in the evaluation dataset. For each instance, the model generates a ranked list of predictions, typically consisting of bounding boxes or class labels. The Acc@1 metric measures the percentage of instances where the ground truth annotation for the referred object aligns with the top-ranked prediction made by the model. 
\textit{IoU@0.5} metric is a commonly used evaluation measure to assess the accuracy of object localization. Formally, let $N$ denote the total number of object referral instances in the evaluation dataset. For each instance, the model generates a predicted bounding box for the referred object, and there is a corresponding ground truth bounding box provided. The IoU@0.5 metric calculates the percentage of instances where the Intersection over Union between the predicted bounding box and the ground truth bounding box exceeds or equals 0.5.

\textbf{Gaze Prediction} involves predicting the spatial probability distribution of a person's gaze within a given visual scene in autonomous driving. Commonly used evaluation metrics include Pearson’s Correlation Coefficient (CC)~\cite{pearson1895note}, Kullback–Leibler Divergence (KLD)~\cite{kullback1951information}, Information Gain (IG)~\cite{quinlan1986induction}, Normalized Scanpath Saliency (NSS), and Similarity Metric (SIM). To be specific, 
\textit{CC}~\cite{pearson1895note} measures the linear relationship between two variables. It quantifies the strength and direction of the linear association between the predicted attention map and the ground-truth fixations. Pearson's correlation coefficient ranges from -1 to 1, where a value of 1 indicates a perfect positive linear relationship, 0 indicates no linear relationship, and -1 indicates a perfect negative linear relationship.
\textit{KLD}~\cite{kullback1951information} quantifies the amount of information lost when comparing the probability distribution of the predicted attention maps to the ground-truth distribution. A smaller KLD value indicates a lower amount of information loss, meaning that the predicted maps closely resemble the ground-truth distribution.
\textit{NSS} calculates the mean value of positive positions in the predicted attention map. It measures how well the predicted attention map aligns with the ground-truth fixations, with higher values indicating better alignment.
\textit{SIM} evaluates the similarity between the predicted attention map and the ground-truth distribution. A larger SIM value indicates a better approximation of the ground-truth distribution by the predicted attention map.

\textbf{Question Answering}. In autonomous driving scenarios, the Question Answering task refers to the process of answering questions related to the visual perception of the autonomous vehicle. It involves analyzing the visual data captured by the vehicle's sensors, such as cameras, LiDAR, or radar, and providing meaningful answers to questions about the environment. The questions in NuScenes-QA~\cite{nuscenes-qa} can be categorized into five groups based on their query formats. The first category is ``Exist'', which involves querying whether a particular object exists in the scene. The second category is ``Count'', where the model is asked to count objects in the scene that meet specific conditions mentioned in the question. The third category is ``Object'', which tests the model's ability to recognize objects in the scene based on language descriptions. The fourth category is ``Status'', which involves querying the status of a specified object. Lastly, the fifth category is ``Comparison'', where the model is requested to compare specified objects or their statuses.

\begin{figure*}[tbp]
    \centering
    \includegraphics[width=0.83\textwidth]{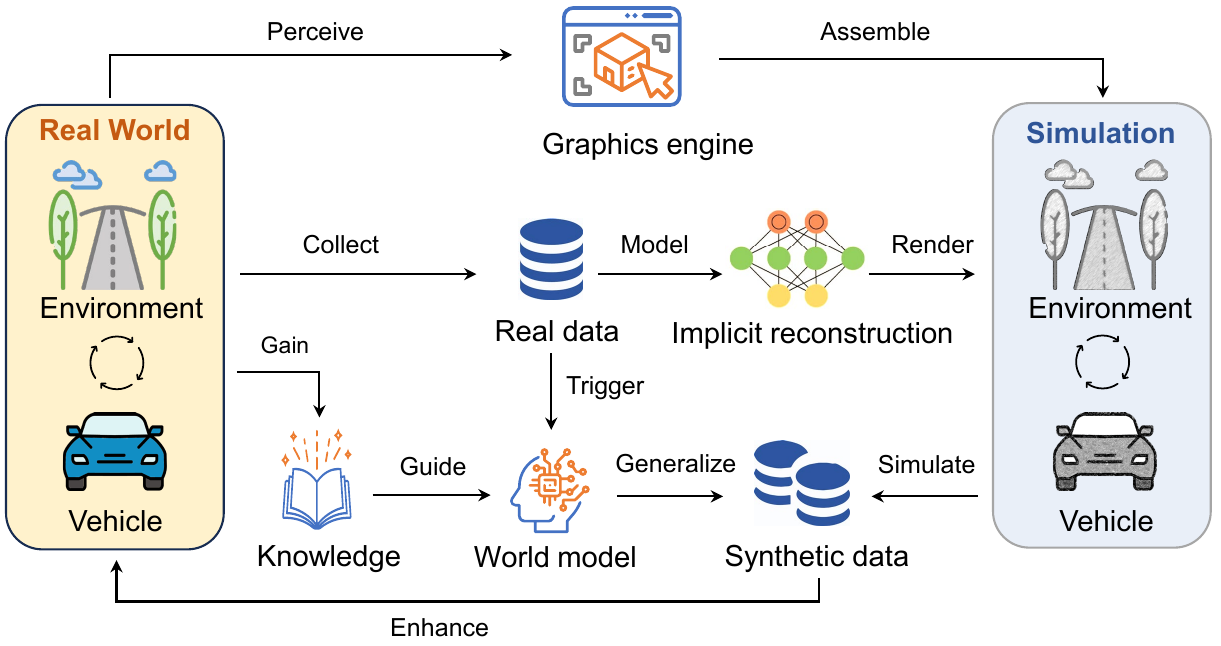}
    \caption{From the real-world environment to the virtual simulation environment. The utilization of graphics engines enables the perception of real-world environments and the assembly of virtual simulated environments, while this approach incurs high costs. Implicit reconstruction methods, which render simulated environments by collecting data from multiple sources, emerge as a promising and cost-effective solution. Integrating knowledge and data to construct world models facilitates a genuine understanding of the environment, enabling the accomplishment of diverse tasks, particularly in synthesizing data to support closed-loop simulations.} 
    \label{fig_env}
\end{figure*}
\section{Environment}

Similar to other AI agent systems, autonomous driving systems require continuous iteration through training to enhance performance, thereby strengthening their adaptability in the environment. Training can utilize collected datasets from real-world environments or take place within constructed closed-loop simulation environments \cite{stocco2022mind, zhang2022rethinking, feng2023dense}. Previous autonomous driving algorithms predominantly rely on the former approach. This involves initial offline training and testing using collected data, followed by deploying the model on vehicles for on-road testing. New issues identified during road testing prompt engineers to repeat the entire process. However, this process involves considerable human and material resources, as 
several stages incur significant costs, including data collection, annotation, and model training \cite{yu2022autonomous}. Thus, some researchers have shifted focus towards ``virtual testing'' \cite{li2019parallel, stocco2022mind}. Shadow mode \cite{othman2022exploring} represents a typical virtual testing approach to self-supervised training by constructing supervisory signals based on the real environment and human driver decisions. Shadow mode enables cloud-based training through data feedback or on-vehicle training through federated learning. Testing on simulation engines is another highly anticipated approach \cite{dosovitskiy2017carla, lopez2018microscopic, wen2023limsim}. The self-training and iteration processes within simulated environments can reduce the costs of data collection and annotation and more closely with human learning skills: observation, interaction, and imitation \cite{zador2022toward}. This methodology is expected to play a crucial role in the knowledge-driven autonomous driving paradigm. Additionally, the emergence of world models enables us to contemplate key issues in scene understanding and construction from the perspective of generative models. As shown in Fig. \ref{fig_env}, we demonstrate the combination of the real-world environment and the virtual simulation.

This section shows the role of the environment in knowledge-driven autonomous driving from three perspectives: (1) simulation engines; (2) high-fidelity sensor simulation; (3) world models.

\subsection{Simulation Engines}
\label{sec:simulationengine}
Simulation engines for autonomous vehicles refer to computer-based simulations of real-world scenarios, including urban roads, highways, various weather conditions, and traffic situations, to facilitate improved training and evaluation of algorithm performance \cite{zhao2023development}. Compared to traditional on-road testing, simulation engines offer several advantages. Firstly, simulation engines provide a safer and more controlled environment, mitigating potential risks associated with testing autonomous vehicles on real roads. Secondly, simulation engines can generate large-scale annotated datasets, which are crucial for training deep learning models. Additionally, simulation engines assist development teams in faster iteration and debugging, enabling anomaly detection and algorithm optimization, thereby enhancing development efficiency. Lastly, simulation engines can generate diverse scenarios in a well-controlled environment, ensuring that the system can respond correctly to various challenges.

In existing autonomous driving systems, distinct simulation tools are available for each stage, including perception, decision-making, planning and control. For example, SUMO \cite{lopez2018microscopic} and LimSim \cite{wen2023limsim} can simulate traffic flows and model the motion interactions between vehicles; HighwayEnv \cite{highway-env}, nuPlan \cite{caesar2021nuplan}, and waymax \cite{waymax} provide closed-loop simulation for decision-making; CarSim \cite{sayers1999vehicle} provides simulation for vehicle dynamics. However, comprehensive testing for autonomous driving necessitates simulation engines that encompass various stages, creating a simulation environment closely resembling the real world. Therefore, Virtual Test Drive \cite{vtd} and CARLA \cite{dosovitskiy2017carla} are designed based on game engines, such as Unreal Engine \cite{Unreal} and Unity Engine \cite{Unity}, aiming to establish three-dimensional end-to-end closed-loop simulation environments. Nevertheless, this construction method still demands substantial human and material resources for manually designing road structures and creating three-dimensional objects, posing challenges for large-scale applications. Moreover, simulators based on these game engines still exhibit significant deficiencies, contributing to domain gaps that impair the accuracy of algorithms trained in simulated scenarios when applied in the real world.

\begin{figure*}[tbp]
    \centering
    \includegraphics[width=1\textwidth]{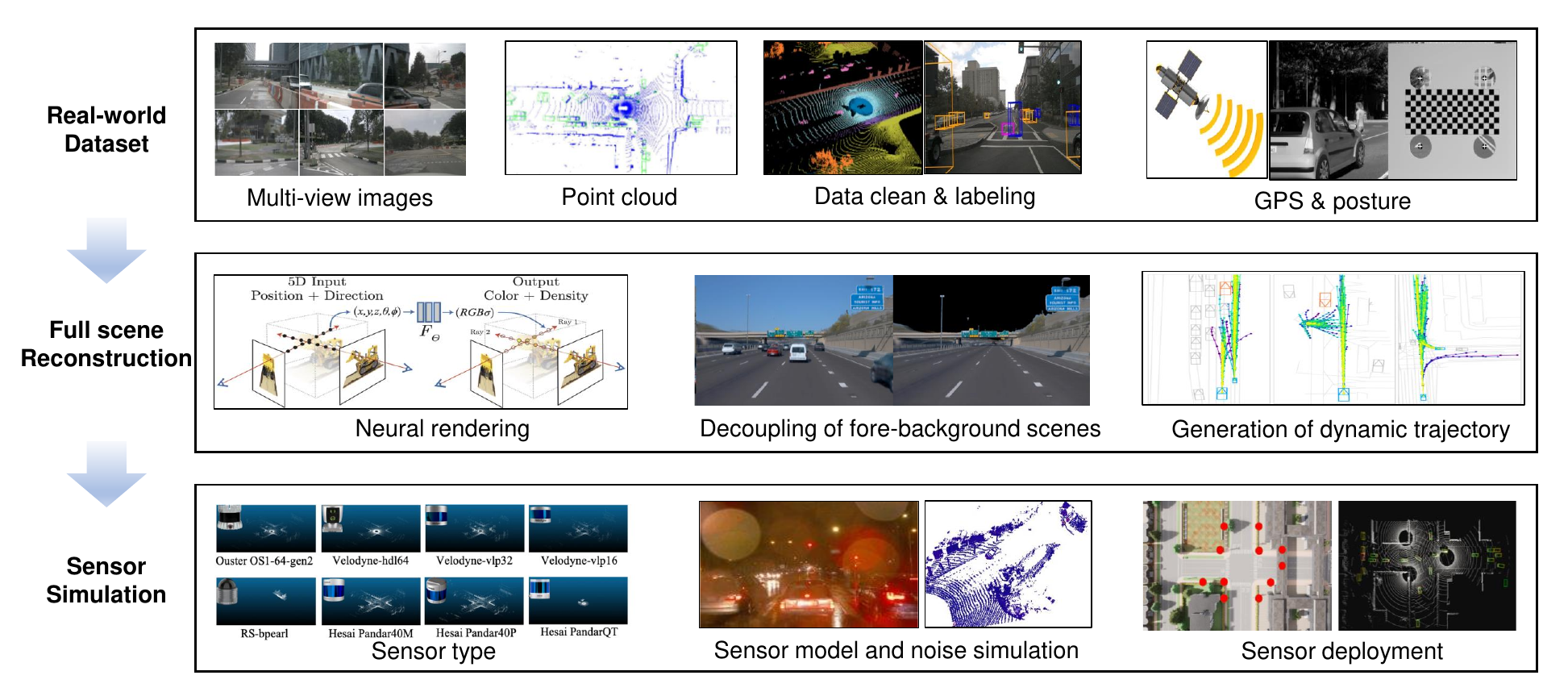}
    \caption{Generalized environment based on neural rendering. (1) Real-world data: Processing and annotating multi-view images and point clouds, and comprehending scenes through information derived from GPS and poses; (2) Full scene reconstruction: Neural rendering technology can decouple and reconstruct foreground and background separately, and various generalized scenes can be generated using dynamic trajectory generation technology; (3) Sensor simulation: Exploring different types of in vehicle sensors, different layout schemes, and simulations under weather and other disturbances.
    } 
    \label{fig_sim}
\end{figure*}

The current research trend involves integrating real-world data into simulation engines \cite{li2022metadrive}. Firstly, the realism of simulation engines can be heightened by leveraging knowledge gained from real-world data. Furthermore, datasets often contain precise annotation information, facilitating a more comprehensive evaluation of the capabilities of the autonomous driving algorithm across different perception and decision-making modules within the simulation engine. Lastly, real-world data may encompass challenging scenarios, and incorporating these scenarios into simulation engines aids in testing the algorithms' robustness when confronted with various challenges. It is noteworthy that, as datasets cannot cover all conceivable scenarios, simulation engines also need to synthesize data to encompass diverse scenarios.

\subsection{High-Fidelity Sensor Simulation}

Although synthesizing large-scale data through simulation engines is advantageous for training autonomous driving systems \cite{li2019aads}, achieving high-fidelity sensor simulation within these engines is a current research challenge. Due to the poor rendering realism of autonomous driving simulators based on game engines \cite{yang2020surfelgan}, meeting the requirements of end-to-end closed-loop simulation for sensor simulation becomes difficult. Consequently, models trained in these closed-loop simulators struggle to reflect their real-world capabilities. Rendering quality has thus become a focal area of research in simulation engine technology.

In recent years, the emergence of neural rendering technologies has shed light on this direction, like Neural Radiance Fields (NeRF) \cite{mildenhall2021nerf}. Neural rendering models objects through implicit representations, calculating the difference between the rendering result and ground truth and using backpropagation to refine the representation, ultimately achieving high-quality 3D reconstruction and rendering. Following the introduction of neural rendering technology, it rapidly expanded from single-object reconstruction to applications in indoor environments \cite{chen2023structnerf, wu2022scalable, chang2023depth, wei2023depth}, static scenes (BlockNeRF \cite{tancik2022block}), and dynamic scenarios (NeuRAD \cite{Tonderski2023NeuRAD}).

Subsequently, UniSim \cite{yang2023unisim} achieved decoupled 3D reconstruction of foreground objects, demonstrating generalization and the ability to generate new data. StreetSurf \cite{guo2023streetsurf} achieved decoupled reconstruction of close-range, mid-range (streets), and far-range (sky) scenes, further enhancing the quality of street scene reconstruction. MARS \cite{wu2023mars} also utilized NeRF technology to construct an autonomous driving simulation engine. Additionally, ReSimAD \cite{zhang2023resimad} validated the performance improvement brought about by applying data generated by neural rendering to perception algorithm training, demonstrating the importance of high-fidelity sensor simulation.

Despite the widespread attention neural rendering technology has garnered in academia and industry, and ongoing efforts to better apply this technology to autonomous driving scenarios, challenges persist in constructing simulation engines based on neural rendering technology. Firstly, neural rendering technology fundamentally remains a 3D reconstruction algorithm, demanding high-quality reconstruction data and sensitivity to motion blur, pose errors, lighting changes, lens flares, and other factors in input data. Secondly, the pursuit of high-fidelity in 3D reconstruction compromises its generalization, making it challenging to generate photorealistic virtual scenes like diffusion models \cite{yang2022diffusion}, GANs \cite{goodfellow2014generative}, and other generative models. Thirdly, large-scale scene reconstruction and rendering pose significant computational challenges, impacting the feasibility of constructing sensor-level high-fidelity simulation engines using neural rendering in terms of reconstruction speed and real-time rendering.

Due to the limited generalization capabilities of neural rendering for scenes, using it for environment simulation can only originate from reconstruction data, making it difficult to contribute to the generation of corner cases that are relevant to autonomous driving simulation. Fig.~\ref{fig_sim} demonstrates a promising technical framework for the generalized environment based on neural rendering. Drawing upon multi-view images, LiDAR-collected point clouds, precise GPS coordinates, sensor pose, and multi-sensor calibrations \cite{yan2022opencalib}, neural rendering technology exhibits the capability to independently reconstruct the foreground and background within a given scene. The foreground reconstruction encapsulates the nuanced portrayal of movements and interactions among traffic participants. The latest dynamic trajectory generation techniques \cite{jiang2023motiondiffuser, zhong2023guided} can facilitate the generation of varied traffic flows distinct from the original scene. The achievement of high-fidelity sensor simulation necessitates a thorough consideration of diverse sensor types, placements \cite{cai2023analyzing}, and potential environmental perturbations, including those induced by varying weather conditions.
\begin{table*}[tbp]
\centering
\begin{threeparttable}
\caption{Overall comparison with existing methods to generate realistic driving scenarios. }
\label{tab:worldmodels}
\renewcommand{\arraystretch}{1.3}
\begin{tabular}{p{2.8cm}p{0.5cm}<\centering|p{1.2cm}<\centering p{1.2cm}<\centering p{1.2cm}<\centering p{1.2cm}<\centering p{1cm} <\centering p{1cm}<\centering}
\toprule[2pt]
\multicolumn{2}{c|}{\multirow{2}{*}{Method}} & \multicolumn{2}{c|}{Priors} & \multicolumn{2}{c|}{\begin{tabular}[c]{@{}c@{}}Outputs\end{tabular}} & \multicolumn{2}{c}{\begin{tabular}[c]{@{}c@{}}Metrics$^\dagger$\end{tabular}} \\ \cline{3-8} 
\multicolumn{2}{c|}{} & \multicolumn{1}{c|}{Box} & \multicolumn{1}{c|}{HD Map} & \multicolumn{1}{c|}{Mutil-view} & \multicolumn{1}{c|}{Video} & \multicolumn{1}{c}{FID$\downarrow$} & \multicolumn{1}{c}{FVD$\downarrow$} \\
\midrule[1pt]
BEVGen~\cite{swerdlow2023street} & & \ding{55} & \ding{51} & \ding{51} & & 25.54 & - \\
BEVControl~\cite{yang2023bevcontrol} & & \ding{55} & \ding{51} & \ding{51} & & 24.85 & - \\
MagicDrive~\cite{gao2023magicdrive} & & \ding{51} & \ding{51} & \ding{51} & & 16.20 & - \\
DrivingDiffusion~\cite{li2023drivingdiffusion} & & \ding{55} & \ding{51} & \ding{51} & & 15.85 & - \\
WoVoGen~\cite{lu2023wovogen} & & \ding{51} & \ding{51} & \ding{51} &  & 27.60 &-\\
Drive-WM~\cite{wang2023driving} & & \ding{51} & \ding{51} & \ding{51} & & 12.99 & - \\ \midrule[1pt]
GAIA-1~\cite{hu2023gaia} & & \ding{51} & \ding{51} & & \ding{51} & - & - \\
% DriveGAN~\cite{kim2021drivegan} & & \ding{55} & \ding{55} & & \ding{51} & 73.4 & 502.3 \\
DriveDreamer~\cite{wang2023drivedreamer} & & \ding{51} & \ding{51} & & \ding{51} & 52.6 & 452.0 \\
DrivingDiffusion~\cite{li2023drivingdiffusion} & & \ding{51} & \ding{51} & & \ding{51} & 15.8 & 332.0 \\
Drive-WM~\cite{wang2023driving} & & \ding{51} & \ding{51} & & \ding{51} & 15.8 & 122.7 \\
WoVoGen~\cite{lu2023wovogen} & & \ding{51} & \ding{51} & & \ding{51} & - & 417.7 \\
ADriver-I~\cite{jia2023adriver} & & \ding{55} & \ding{55} & & \ding{51} & {5.5} & {97.0} \\
\bottomrule[2pt]
\end{tabular}
\begin{tablenotes}
\item \small {$\dagger$ means that the generation quality are evaluated on the nuScenes~\cite{nuscenes}.}
\end{tablenotes}
\end{threeparttable}
\vspace{-2pt}
\end{table*}

\subsection{Environment Understanding by World Model}

The world model aims to simulate and understand physical laws and phenomena in the real world, or can be considered as an abstract representation of the environment \cite{ha2018world}. The main idea of the model is to build an abstract representation of the real world by learning data acquired from multiple sensors, such as images, sounds and sensor data. The model can then use this abstract representation to make inferences and predictions in order to make decisions about unknown situations. Such models have a wide range of applications in areas such as robot control, autonomous driving, and game AI. Currently world model is usually built as an end-to-end deep learning framework that can train using self-supervised or weakly supervised methods directly from raw sensor data without extracting features manually. The advantage of this model is that it can handle complex nonlinear relationships between different objects in the scene and adaptively fit different environments and tasks.  This makes the world model a universal way of understanding the real world, similar to the way humans think \cite{mitchell2023ai}. The JEPA model \cite{lecun2022path} aims to construct mapping relationships between different inputs in the encoding space by minimizing input information and prediction errors. The world model can enhance the ability of autonomous driving to understand the environment and support large-scale high-quality driving video generation~\cite{yang2023bevcontrol,gao2023magicdrive,jia2023adriver,li2023drivingdiffusion,lu2023wovogen}, as shown in Table~\ref{tab:worldmodels}. For example, DriveDreamer\cite{wang2023drivedreamer} uses a diffusion model to construct a comprehensive representation of complex environments, enabling recognition of structured traffic constraints and the ability to predict the future. GAIA-1 \cite{hu2023gaia} is a fully end-to-end generative model that utilizes video, text, and action inputs to generate real driving scenarios, and also enables prediction of future tokenized sequences. Differing from the aforementioned approaches, Zhang et al. \cite{zhang2023learning} propose an unsupervised world model on sensor data derived from point clouds, it tokenizes point clouds using a vector quantized variational autoencoder (VQVAE)~\cite{van2017neural} combined with a PointNet and adopts a combination of generative masked modeling and discrete diffusion for learning a world model. OccWorld~\cite{zheng2023occworld} can forecast future scene evolutions and ego movements jointly based on the given past 3D occupancy observations in a self-supervised manner.

The predictive capability of world models involves inferring the relative positions and movement trends of other vehicles based on current and past scene information, enabling the modeling of potential effects of various actions and informed decision-making \cite{zhang2023trafficbots, zhang2023learning}. Beyond merely predicting original sensor signals, world models are intended to emulate human thinking and comprehension of the real world. To achieve this, world models need to incorporate expert experience embedding and interactive learning \cite{martino2023knowledge, lenat2023getting, agrawal2023can}, enhancing their multitasking capabilities and establishing them as foundational models for knowledge-driven autonomous driving \cite{chen2023driving}.

\section{Driver Agents}
The section initially delves into the development of embodied AI and its connection to autonomous driving. Following that, it succinctly summarizes recent studies focusing on LLMs in autonomous driving, leveraging their robust reasoning and interpretable capabilities. Ultimately, a generalized knowledge-driven framework is introduced, spotlighting crucial components such as cognition, memory, planning, and reflection, with the overarching goal of enhancing scene understanding and decision-making.

\subsection{Embodied AI}
Embodied AI~\cite{pfeifer2004embodied, smith2005development, surveyembodiedai} is a facet of intelligence emphasizing the direct interaction between an intelligent system and its environment, involving perception, understanding, and action. Notably, advancements in embodied intelligence have concentrated on humanoid robots and embodied AGI. As the ideal form of embodied AI, humanoid robots have been improving their autonomy, flexibility and intelligence \cite{zhu2023ghost}, such as the Optimus humanoid robot introduced by Tesla, whose motion control ability has been evolving, providing a strong hardware foundation for the development of embodied AI. Meanwhile, embodied AGI is also considered an important way to realize advanced AI and has attracted the attention of many scholars \cite{law2023artificial}.

LLMs are anticipated to elevate natural and human-like text and image interactions within the domain of embodied AI \cite{driess2023palme}. They play a pivotal role in assisting embodied AI systems in comprehending and perceiving their surroundings, interpreting intricate task descriptions, formulating task plans, collaborating seamlessly with other system modules, adapting to dynamic environments, and facilitating social interactions with humans through natural language exchanges \cite{wang2023survey, xi2023rise}. Despite these advantages, it is imperative to address potential drawbacks, such as decision uncertainty. The uncertainties of LLMs also bring risks to embodied AI, potentially resulting in biases or errors in information processing, thereby compromising the systems' functionality and the reliability of task completion.

Autonomous driving can be considered within the realm of embodied AI, whereas the open and dynamic traffic environment faced by autonomous driving necessitates a heightened focus on system reliability and generalization~\cite{oravec2022future}. While autonomous driving can rely on the common sense understanding and logical reasoning ability of LLMs, they cannot completely rely on LLMs' output as final decisions. Therefore, adopting a knowledge-driven paradigm can enhance autonomous driving by integrating mechanisms for long-term learning and knowledge accumulation, facilitating prompt adaptation to environmental changes through immediate feedback and adjustment.
\begin{table*}
% \footnotesize
\centering
\caption{Knowledge-driven methods based on LLMs in Autonomous Driving.}
\label{tab:llmad}
\renewcommand{\arraystretch}{1.5}
\begin{tabular}{c|m{3.2cm}|m{3cm}|m{8.3cm}}
\toprule[2pt]

\multicolumn{1}{c|}{\centering\textbf{Category}} & \multicolumn{1}{c|}{\centering\textbf{Methods}} & \multicolumn{1}{c|}{\centering\textbf{Modalities}} & \multicolumn{1}{c}{\centering\textbf{Characteristics}} \\
\bottomrule[1pt]
\multirow{2}{*}[-1.5ex]{Perception} 
& Language Prompt~\cite{wu2023language}& Image, Text & LLM (GPT-3.5~\cite{chatgpt}), language prompts, tracking \\ \cline{2-4}
& \makecell[l]{Can You Text What \\ Is Happening}~\cite{keysan2023can} & Image, Text & LLM (DistilBERT~\cite{sanh2019distilbert}), trajectory prediction \\
\midrule[1pt]
\multirow{5}{*}[-8ex]{\shortstack{  Decision-making \\ \& \\ Planning  \\ \&  \\ Control}} 
& Drive Like A Human~\cite{fu2023drive} & 2D BEV, Text & LLM (GPT-3.5), closed-loop system, decision-making and control. \\ \cline{2-4}
& Drive as You Speak~\cite{cui2023drive} & 2D BEV, Map, GNSS, Radar, LiDAR, Image & LLM (GPT-4~\cite{openai2023gpt4}), decision-making \\ \cline{2-4}
& DiLu~\cite{wen2023dilu} & Text & LLM (GPT-3.5), agent, memory module, knowledge, reasoning, decision-making and control \\ \cline{2-4}
& LanguageMPC~\cite{sha2023languagempc} & Text & LLM (GPT-3.5), decision-making and control\\\cline{2-4}
& Talk2BEV~\cite{dewangan2023talk2bev}& Image, Text & large vision language model (LVLM) BLIP-2~\cite{li2023blip}
 and LLaVA~\cite{liu2023visual}, augmented bird’s-eye view (BEV) maps \\ \cline{2-4}
& TrafficGPT~\cite{zhang2023trafficgpt} & Text & LLM (GPT-3.5), analyze, decision-making \\\cline{2-4}
& Receive Reason React~\cite{cui2023receive} & Text & LLM (GPT-4), reasoning, decision-making \\ 
\midrule[1pt]
\multirow{5}{*}{End-to-End} 
& DriveGPT4~\cite{xu2023drivegpt4} & Image, Text, Action &  LLM (LLaMA2~\cite{touvron2023llama}), action, reasoning. \\ \cline{2-4}
& GPT-Driver~\cite{mao2023gpt} & Image, Text, Action & LLM (GPT-3.5), motion planner, trajectory generation and control \\ \cline{2-4}
& Drive Any where~\cite{wang2023drive} & Image, Text & LLM (BLIP) open set learning,
ViT~\cite{dosovitskiy2020image}, perception,\\ \cline{2-4}
& Agent-Driver~\cite{mao2023language} & Image, Text, Action & LLM (GPT-3.5), agent, tool library, reasoning, cognitive memory \\ \cline{2-4}
& DESIGN-Agent~\cite{anonymous2023d} & Image, Text & LLM (GPT-3.5), agent, reasoning\\
\midrule[1pt]
\multirow{3}{*}{\shortstack{VQA \\ \& \\ Captioning}} 
& Driving with LLMs~\cite{chen2023driving} &Text  & LLM (LLaMa~\cite{touvron2023llama}, GPT-3.5) questions answering \\ \cline{2-4}
& Dolphins~\cite{ma2023dolphins} & Image, Text & LLM (OpenFlamingo~\cite{awadalla2023openflamingo}), Vision Language Action, Grounded Chain of Thought (GCoT), reflection\\ \cline{2-4}
& LINGO-1~\cite{Wayve-LINGO-1} & Image, Text, Action & LLM (GPT-3.5), Vision Language Action, reasoning\\
\midrule[1pt]
\multirow{1}{*}{Simulation} 
& SurrealDriver~\cite{jin2023surrealdriver} & Text & LLM (GPT-3.5), generative simulation, human-like driving behaviors\\
\bottomrule[2pt]
\end{tabular}
\end{table*}

\subsection{Applying LLMs to Enhance Autonomous Driving}

As shown in Table~\ref{tab:llmad}, with the rapid advancement of LLMs, provides a foundation for injecting human knowledge and common sense into Driver Agents, sparking numerous new research endeavors. This learning ability is of significance in the \textbf{perception} module of the
autonomous driving system, which greatly improves the system’s adaptability and generalization capabilities in changing and complex driving environments. Talk2BEV~\cite{dewangan2023talk2bev} augment BEV maps with language to enable general-purpose linguistic reasoning for driving scenarios. LanguagePrompt~\cite{wu2023language} uses language prompts as
semantic cues and combines LLMs with 3D detection tasks
and tracking tasks. Although it achieves better performance
compared to other methods, the advantages of LLMs do not
directly affect the tracking task. Rather, the tracking task
serves as a query to assist LLMs in performing 3D detection tasks. As for \textbf{planning, decision-making and control} in autonomous driving, numerous studies aim to harness the robust common-sense comprehension and reasoning capabilities of LLMs to aid drivers~\cite{fu2023drive,cui2023drive}. Some works seek to emulate and even fully replace drivers~\cite{wen2023dilu,cui2023receive,xu2023drivegpt4,sha2023languagempc}. When employing LLMs for closed-loop control in autonomous driving, the majority of research efforts~\cite{fu2023drive,wen2023dilu,cui2023receive} incorporate a memory module to capture driving scenarios, experiences, and other crucial driving information. As well known, an \textbf{end-to-end autonomous driving} system takes raw sensor data as input and generates a plan and/or low-level control actions as output. We recognize end-to-end autonomous vehicle aligns seamlessly with the structure in multimodal input-to-text in LLMs.  this inherent compatibility, several studies are now exploring the viability of integrating LLMs into end-to-end autonomous driving. In contrast to conventional end-to-end autonomous driving systems~\cite{hu2023planning,jia2023driveadapter}, end-to-end autonomous driving systems based on LLMs exhibit robust interpretability, trustworthiness, and advanced scene comprehension capabilities, which opens up avenues for practical application and implementation of end-to-end autonomous driving~\cite{mao2023gpt,mao2023language,xu2023drivegpt4,anonymous2023d,wang2023drive}.
\textbf{Understanding} driving scenes like visual question answering or captioning tasks at a correct and high level is crucial for ensuring driving safety. DrivingLLM~\cite{chen2023driving} evaluate the model’s performance in the driving scene
with a visual and spatial understanding based on visual question answering or captioning tasks. More recently, in showcasing the proficiency of GPT-4V~\cite{openai2023gpt4}, On The Road With GPT-4V~\cite{wen2023road} provide comprehensive tests on GPT-4V in both diverse traffic scenarios and span from basic scene understanding to complex causal reasoning. Various exploratory efforts have utilized Vision Language Models (VLMs) to comprehend traffic scenes through specific downstream tasks. As mentioned in Sec ~\ref{sec:simulationengine}, \textbf{simulation} is pivotal in the advancement of autonomous driving. Yet, existing simulation platforms face constraints in replicating the realism and diversity of agent behaviors, hindering the effective translation of simulation results into real-world applications. SurrealDriver~\cite{jin2023surrealdriver} introduces a novel generative driver agent simulation framework, leveraging LLMs. It demonstrates the ability to perceive intricate driving scenarios and generate realistic driving maneuvers.

\begin{figure*}[t]
    \centering
    \includegraphics[width=0.75\textwidth]{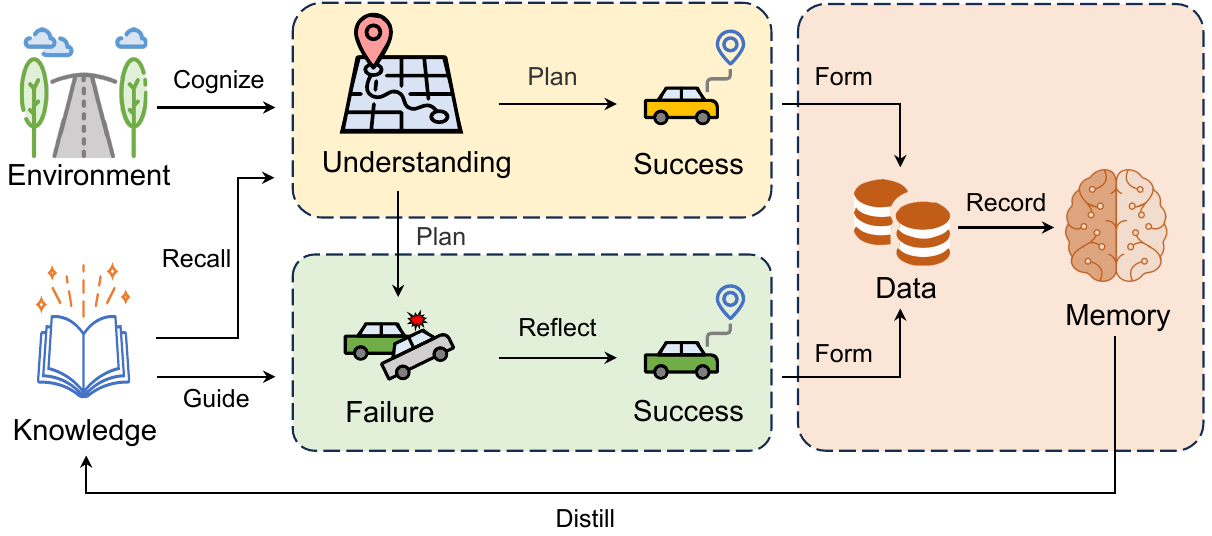}
    \caption{Generalized knowledge-driven framework.} 
    \label{fig_framework}
    \vspace{5pt}
\end{figure*}

The common sense understanding and logical reasoning of LLMs are vital for autonomous driving. However, direct applications of LLMs decision-maker may face challenges~\cite{fu2023drive, cui2023drive}. To address this, adopting few-shot prompts guides the model in understanding unknown scenarios, considering interpretability and reasonability \cite{wen2023dilu}. Despite the advantages of few-shot prompts, challenges exist, especially in complex tasks where the number of prompts may be insufficient. Building powerful autonomous driving systems involves fine-tuning generalized models for specific driving scenarios \cite{peng2023instruction}, leveraging deep learning on extensive driving data. Autonomous driving systems need to comprehensively understand traffic environments, road structures, and human behavior, integrating text and image information for enhanced perception. Incorporating interaction processes and competitive games enables systems to grasp behaviors with other traffic participants and learn complex decision-making strategies. Large-scale training in simulators improves generalization, while iterative optimization, real-time feedback, and emphasis on safety standards lead to continuous improvement in model performance.
% and intelligence.

\subsection{Generalized Knowledge-driven Framework}
A generalized knowledge-driven framework, inspired by recent advancements like Smallville \cite{park2023generative}, Dilu \cite{wen2023dilu}, LLM-Brain \cite{mai2023llm}, etc., is essential for autonomous driving. This framework integrates various components and technologies, as depicted in Fig.~\ref{fig_framework}, encompassing cognition, planning, reflection, memory, and more. Cognitive understanding transcends traditional detection and segmentation tasks, demanding a profound comprehension of specific task environments. Crucially, planning correct actions based on object relationships becomes pivotal, with autonomous reflection necessary in the face of decision failures leading to anomalies. The memory module is enriched by both positive and negative samples, contributing to knowledge distillation. In a closed-loop continuous learning system, accumulated knowledge guides decision-making and reflection processes. Despite the general domain knowledge provided by rapidly advancing LLMs, precise performance in autonomous driving tasks mandates the empowerment and enhancement of knowledge-driven frameworks.

\textbf{Cognition.} Various sensors such as cameras, radar, and LiDAR are employed to capture environmental information, subsequently transformed into semantic representations of the environment \cite{li2020building, berrio2021camera, premebida2013fusing, wang2023lidar2map}. This information can be processed by leveraging LLMs, enabling semantic understanding and logical reasoning \cite{de2023semantic, dewangan2023talk2bev, wu2023next}. LLM-based systems demonstrate the capability to identify objects on roads and comprehend traffic signs \cite{elhafsi2023semantic, zhou2023vision}. However, to enhance scene understanding, LLMs necessitate closed-loop environments, incorporating positive and negative feedback, overcoming illusions, and continuously expanding knowledge through lifelong learning \cite{wang2023voyager, zhao2023expel}. Cognition, involving the comprehension of objects and their interconnections, demands continuous fine-tuning of cognitive models in autonomous driving to address scenarios from simplistic to sophisticated through interaction with the environment.

\textbf{Memory.} The outcomes of semantic understanding are stored in the internal memory, constructing a dynamic perception of the environment \cite{zhang2023memory}. This enables the system to retain and continually update its understanding of the surroundings. Furthermore, historical driving experiences and knowledge are archived in the internal memory. When confronted with a comparable situation, the system retrieves past semantic understanding and driving decisions to adeptly address similar scenarios. Distinctions between long-term and short-term memory are essential. Memory cultivated through numerous similar scenes fine-tunes the foundation model, fostering a rapid reasoning ability akin to human unconditional reflection. Contrasely, short-term memory only preserves recent and unfamiliar scenarios, ensuring swift adaptation to diverse environments.

\textbf{Planning.} By amalgamating sensing results, historical knowledge, and LLMs' reasoning capabilities, the system formulates dections for path planning, speed control, and obstacle avoidance \cite{mao2023gpt, wang2023chatgpt, sha2023languagempc}. Ensuring planned behaviors align with traffic rules and safety standards is crucial for achieving secure autonomous driving. While LLMs serve as a means of knowledge extraction and utilization, they function as a linguistic bridge between existing human knowledge and machine execution processes, facilitating interpretive reasoning and decision-making. However, LLMs, as carriers of general knowledge, require artificially designed prompts and feed shots for application in vehicle manipulation. Moreover, relying solely on LLMs for driving decisions is a transitional approach; developing large-scale symbolic models tailored to autonomous driving represents a more specialized avenue.

\textbf{Reflection.} The driving decisions undergo interpretation using LLMs, contributing to an understanding of systems' behaviors. Analyzing the LLMs' outputs allows for the evaluation of the system's decision rationality, facilitating continuous optimization and learning to enhance performance and robustness \cite{shinn2023reflexion, cui2023drivellm, olausson2023demystifying}. Additionally, reflection can incorporate expert systems, leveraging accident cases from datasets or human-derived lessons to swiftly identify and localize potential issues, thereby finding suitable solutions for knowledge-driven systems.

\section{Opportunities and Challenges}

\textbf{Knowledge embedding dataset.} Ensuring dataset richness involves covering daily driving situations, emergencies, and extreme weather conditions. This diversity enhances the model's ability to understand and adapt to various realistic driving scenarios comprehensively. The use of natural language annotation, closely resembling a driver's thought and decision-making process, improves the model's understanding of human behavior and aligns it with real driving cognition. Annotators with ample driving experience ensure accurate annotation of diverse driving situations, focusing on scenario understanding for enhanced accuracy and quality. While language has demonstrated impressive proficiency in knowledge-embedding datasets, it cannot be conclusively stated that language is the only way to represent knowledge. Therefore, delving into more suitable methods of knowledge representation presents a worthy research direction.

\textbf{Efficient and realistic virtual environment.} Virtual environments need to overcome challenges through refined neural rendering technology, achieving efficiency and realism in simulations. Optimizing 3D reconstruction algorithms strikes a balance between high fidelity and generalization, focusing on adaptability. Diverse and realistic virtual landscapes result from independently reconstructing foreground and background using various data sources. Techniques like Gaussian Splatting~\cite{kerbl20233d} offer efficiency in the handling of large-scale scenes, enabling real-time, high-performance virtual driving environments. Proactive exploration in environment understanding aims to construct intelligent models simulating real-world physical laws. Leveraging data from multiple sensors establishes an abstract representation of the environment. Incorporating such environments enhances training and testing for autonomous driving systems, fostering continuous advancements in the field.
    
\textbf{VLMs.} VLMs offer enhanced integration compared to LLMs, aiming to approach human-level perception and understanding. Crucial for decision-making and behavior planning, VLMs excel in surrounding perception and scene understanding \cite{gpt4v, sammani2022nlx, gao2023llamaadapterv2, huang2023voxposer}. VLMs outperform in traffic scenario understanding by fusing visual and linguistic information, and comprehending complex situations like road signs, traffic signs, and pedestrians. Their multimodal semantic understanding ensures reliable interpretation of traffic participants' states and behaviors, particularly excelling in deep understanding and reasoning in intricate scenes. However, it's essential to note that specialized learning is required for VLMs' 3D spatial understanding and driving skills, presenting a focus for future research and development.

\textbf{Requirements and Validation of Knowledge-Driven Approaches.} Knowledge-driven autonomous driving demands enhanced cognitive and understanding capabilities, necessitating comprehension of common objects and intricate relationships between them based on physical laws and traffic rules. This involves understanding vehicle movements, interactions with other traffic participants, and ensuring maneuvers comply with traffic regulations. Knowledge-driven approaches extend beyond traditional performance metrics, requiring comprehensive validation of the entire process, from scenario understanding to vehicle maneuvering. Such validation enhances system transparency, aligns decision-making processes with intuitive human knowledge, and ultimately strengthens the credibility and safety of autonomous driving systems, reducing the risk of generating hallucinatory decisions \cite{ye2023cognitive, rawte2023survey}.

\section{Conclusion}

Knowledge-driven autonomous driving is the revolutionary paradigm that is promising to break through the current bottlenecks of autonomous driving. It emphasizes life-long learning, iterative revolution, and the integration of multimodal data, promising improved performance, safety, and interpretability in autonomous driving systems. The transition towards knowledge-driven autonomous driving reflects a pivotal evolution in technology development, emphasizing scenario understanding and reasoned decision-making. 
First, we introduce the foundational components of knowledge-driven autonomous driving: Dataset \& Benchmark, Environment, and Driver Agent. These components, especially when synergized with advanced technologies like LLMs, world models, and neural rendering, collectively enhance the intelligence of autonomous systems. This integration facilitates a deeper and more holistic interaction with the driving environment, thereby augmenting the system's overall capabilities. Next, we present a comprehensive knowledge-driven framework for autonomous driving, including critical components like cognition, planning, reflection, and memory, aiming to empower autonomous driving systems with scenario understanding, strategic decision-making, and life-long learning. 
Finally, we also highlight opportunities and challenges in knowledge-driven autonomous driving, including the importance of diverse datasets for comprehensive model training, the incorporation of natural language annotation for alignment with human thought processes, the creation of efficient virtual environments through refining neural rendering and optimizing 3D reconstruction, and the integration of LLMs for decision-making and behavior planning in complex driving scenarios, concluding with an emphasis on the verification measures for autonomous vehicles. Nevertheless, the journey towards fully realizing the potential of knowledge-driven autonomous driving is not devoid of challenges. This paper aims to highlight the significance of adopting knowledge-driven approaches in the evolving landscape of autonomous driving technologies. Our objective is to steer future research and practical applications in the direction of creating more intelligent, adaptable, and robust autonomous driving systems.

\bibliographystyle{IEEEtran}
\bibliography{ref}

\vfill

\end{document}